\documentclass{article}

\usepackage{iclr2027_conference,times}
\usepackage{microtype}
\usepackage{xurl}
\usepackage{algorithm}
\usepackage[noend]{algorithmic}
\usepackage{hyperref}

\usepackage{amsmath,amsfonts,bm}

\def\eqref#1{equation~\ref{#1}}

\def\1{\bm{1}}

\newcommand{\malpha}[1][\alpha]{\ensuremath{M_{#1}}}
\newcommand{\mift}{\ensuremath{M_{\mathrm{IFT}}}}
\newcommand{\dtrain}{\ensuremath{\mathcal{D}_{\mathrm{train}}}}
\newcommand{\dval}{\ensuremath{\mathcal{D}_{\mathrm{cal}}}}
\newcommand{\dcal}{\ensuremath{\mathcal{D}_{\mathrm{cal}}}}
\newcommand{\dtest}{\ensuremath{\mathcal{D}_{\mathrm{test}}}}

\def\rvc{{\mathbf{c}}}

\def\rvr{{\mathbf{r}}}

\def\rvx{{\mathbf{x}}}
\def\rvy{{\mathbf{y}}}

\DeclareMathAlphabet{\mathsfit}{\encodingdefault}{\sfdefault}{m}{sl}
\SetMathAlphabet{\mathsfit}{bold}{\encodingdefault}{\sfdefault}{bx}{n}

\def\gD{{\mathcal{D}}}

\usepackage[x11names]{xcolor}
\usepackage{xcolor}
\usepackage{listings}
\usepackage{fvextra}
\usepackage{float}
\usepackage{needspace}
\usepackage{nicematrix}

\usepackage[utf8]{inputenc}
\usepackage{lipsum}

\usepackage[T1]{fontenc}

\usepackage{etoolbox}
\newbool{includeappendix}
\setbool{includeappendix}{true}

\ifdefined\isoverfull
\else
\fi

\usepackage{subcaption}

\definecolor{my-full-blue}{HTML}{1F77B4}

\definecolor{my-full-orange}{HTML}{FF7F0E}
\definecolor{my-extra-orange}{HTML}{7C4514}

\definecolor{my-full-green}{HTML}{2CA02C}

\definecolor{my-full-red}{HTML}{d62728}

\definecolor{my-full-purple}{HTML}{9467bd}

\colorlet{my-blue}{my-full-blue!30}
\colorlet{my-orange}{my-full-orange!30}
\colorlet{my-green}{my-full-green!90}
\colorlet{my-red}{my-full-red!90}
\colorlet{my-purple}{my-full-purple!30}

\definecolor{mygreen}{HTML}{B5E3B5}
\colorlet{myred}{LightPink1}
\colorlet{myblue}{SteelBlue2}
\definecolor{mylightblue}{HTML}{B0DBFF}
\colorlet{myorange}{DarkOrange1}

\definecolor{darkblue}{RGB}{34,49,63}
\definecolor{lightblue}{RGB}{238,244,249}
\definecolor{accentblue}{RGB}{88,139,202}
\definecolor{lightgreen}{RGB}{220,245,230}
\definecolor{lightred}{RGB}{250,225,225}
\definecolor{darkgreen}{RGB}{40,167,69}
\definecolor{darkred}{RGB}{220,53,69}
\definecolor{textgray}{RGB}{120,120,120}
\definecolor{lightgray}{RGB}{240,240,240}
\definecolor{accentorange}{RGB}{127,17,144}
\definecolor{gray2}{HTML}{FCFCFC}

\usepackage{listings}

\usepackage{textcomp}

\usepackage[scaled=0.8]{beramono}

\definecolor{ckeyword}{HTML}{7F0055}
\definecolor{ccomment}{HTML}{3F7F5F}
\definecolor{cstring}{HTML}{2A0099}

\lstdefinestyle{numbers}{
	numbers=left,
	framexleftmargin=20pt,
	numberstyle=\tiny,
	firstnumber=auto,
	numbersep=1em,
	xleftmargin=2em
}

\lstdefinestyle{layout}{
	frame=none,
	captionpos=b,
}

\lstdefinestyle{comment-style}{
	morecomment=[l]//,
	morecomment=[s]{/*}{*/},
	commentstyle={\color{ccomment}\itshape},
}

\lstdefinestyle{string-style}{
	showstringspaces=false,%
}

\lstdefinestyle{keyword-style}{
	keywordstyle={\ttfamily\bfseries},
	morekeywords={
		function,
		constructor,
		int,
		bool,
		return,
		returns,
		uint
	},
	morekeywords = [2]{},
	keywordstyle = [2]{\text},
	sensitive=true,
}

\lstdefinestyle{input-encoding}{
	inputencoding=utf8,
	extendedchars=true,
	literate=
	{ℝ}{$\reals$}1%
	{→}{$\rightarrow$}1%
	{α}{$\alpha$}1%
	{β}{$\beta$}1%
	{λ}{$\lambda$}1%
	{θ}{$\theta$}1%
	{ϕ}{$\phi$}1%
}

\lstdefinestyle{escaping}{
	escapeinside={(*@}{@*)},
	mathescape=true
}

\lstdefinestyle{default-style}{
	basicstyle=\fontencoding{T1}\ttfamily\footnotesize,
	style=numbers,
	style=layout,
	style=comment-style,
	style=string-style,
	style=keyword-style,
	style=input-encoding,
	style=escaping,
	tabsize=2,
	upquote=true
}

\lstdefinelanguage{BASIC}{
	language=C++,
	style=default-style
}[keywords,comments,strings]%

\lstdefinelanguage{JavaScript}{
morekeywords=[1]{break, continue, delete, else, for, function, if, in,
new, return, this, typeof, var, void, while, with, const, let},
morekeywords=[2]{false, null, true, boolean, number, undefined, string,
Array, Boolean, Date, Math, Number, String, Object},
morekeywords=[3]{eval, parseFloat, escape, unescape},
sensitive,
morecomment=[s]{/*}{*/},
morecomment=[l]//,
morecomment=[s]{/**}{*/},
morestring=[b]',
morestring=[b]"
}[keywords, comments, strings]
\definecolor{delim}{RGB}{20,105,176}
\definecolor{numb}{RGB}{106, 109, 32}
\definecolor{string}{rgb}{0.64,0.08,0.08}

\lstdefinelanguage{json}{
    showspaces=false,
    showtabs=false,
    breaklines=true,
    postbreak=\raisebox{0ex}[0ex][0ex]{\ensuremath{\color{gray}\hookrightarrow\space}},
    upquote=true,
    morestring=[b]",
    morecomment=[l]//,
    stringstyle=\color{string},
    literate=
     *{0}{{{\color{numb}0}}}{1}
      {1}{{{\color{numb}1}}}{1}
      {2}{{{\color{numb}2}}}{1}
      {3}{{{\color{numb}3}}}{1}
      {4}{{{\color{numb}4}}}{1}
      {5}{{{\color{numb}5}}}{1}
      {6}{{{\color{numb}6}}}{1}
      {7}{{{\color{numb}7}}}{1}
      {8}{{{\color{numb}8}}}{1}
      {9}{{{\color{numb}9}}}{1}
      {\{}{{{\color{delim}{\{}}}}{1}
      {\}}{{{\color{delim}{\}}}}}{1}
      {[}{{{\color{delim}{[}}}}{1}
      {]}{{{\color{delim}{]}}}}{1},
}
\definecolor{dkgreen}{rgb}{0,0.6,0}
\definecolor{dred}{rgb}{0.545,0,0}
\definecolor{dblue}{rgb}{0,0,0.545}
\definecolor{lgrey}{rgb}{0.9,0.9,0.9}
\definecolor{gray}{rgb}{0.4,0.4,0.4}
\definecolor{darkblue}{rgb}{0.0,0.0,0.6}
\lstdefinelanguage{cpp}{
      breaklines=true,               
      postbreak=\raisebox{0ex}[0ex][0ex]{\ensuremath{\color{gray}\hookrightarrow\space}},
      deletekeywords={...},          
      escapeinside={\%*}{*)},                  
      language=C++,                
      keywordstyle=\color{purple},  
      morekeywords={string,float}, 
      identifierstyle=\color{black},
      stringstyle=\color{blue},      
      showspaces=false,               
      showstringspaces=false,        
      showtabs=false,                
      tabsize=5,                     
    }

\usepackage[english]{babel}
\definecolor{darkpastelblue}{HTML}{0279AF}
\hypersetup{breaklinks=true,
      colorlinks,
      linkcolor=darkpastelblue,
      citecolor=darkpastelblue,
      urlcolor=darkpastelblue,
      filecolor=darkpastelblue}
\usepackage{color,soul}
\usepackage{ bbold }
\usepackage{multirow}
\usepackage{multicol}
\usepackage{caption}
\usepackage{tabularx,booktabs,xltabular,colortbl}
\usepackage{graphicx}
\usepackage{tabu}
\usepackage{array}
\usepackage{siunitx}
\usepackage{subcaption}
\usepackage{fontawesome}
\usepackage{stackengine}
\usepackage{svg}
\usepackage{mathtools}
\usepackage{xspace}
\usepackage{wrapfig}
\usepackage{makecell}
\usepackage{placeins}
\usepackage{float}
\usepackage{pifont}
\usepackage{svg}
\usepackage[most]{tcolorbox}
\usepackage{array}
\usepackage{dsfont}
\usepackage{enumitem}
\usepackage[utf8]{inputenc}
\usepackage{textcomp}
\usepackage{amsthm}
\usepackage{amssymb,amsmath}

\newcolumntype{x}[2]{S[table-format=#1.#2,table-auto-round]}

\usepackage{tikz}

\usetikzlibrary{arrows}
\usetikzlibrary{automata}
\usetikzlibrary{calc}
\usetikzlibrary{backgrounds}
\usetikzlibrary{decorations.markings}
\usetikzlibrary{decorations.pathmorphing}
\usetikzlibrary{decorations.pathreplacing}
\usetikzlibrary{fit}
\usetikzlibrary{patterns}
\usetikzlibrary{positioning}
\usetikzlibrary{shadows}
\usetikzlibrary{shapes}
\usetikzlibrary{shapes.geometric}
\usetikzlibrary{arrows.meta}
\usetikzlibrary{shadows.blur}

\definecolor{blue}{HTML}{347bc6}
\definecolor{green-underline}{HTML}{2de12c}
\definecolor{yellow-underline}{HTML}{ffd700}

\tikzstyle{block} = [
    minimum width=1cm,
    minimum height=0.5cm,
    align=center,
    fill=gray!30,
    rounded corners=5pt,
]

\tikzstyle{arrow} = [
	-{Latex[length=1.4mm, width=1.4mm]},
	draw=blue,
]

\tikzstyle{dashedline} = [
    draw=blue,
    dashed
]

\tikzstyle{line} = [
    draw=blue
]

\definecolor{lightblue}{RGB}{173,216,230}

\usepackage{pgfplots}
\usepackage[capitalize,noabbrev]{cleveref}
\AtBeginDocument{
\crefname{appendix}{Appendix}{Appendices}
}

\usepackage{pifont}

\definecolor{pastelgreen}{HTML}{059C05}
\definecolor{pastelred}{HTML}{FF7373}
\definecolor{mergehighlightblue}{HTML}{EAF4FF}
\definecolor{iftreasoningred}{HTML}{FFE5E5}

\usepackage{setspace}

\renewcommand{\algorithmicrequire}{\textbf{Input:}}
\renewcommand{\algorithmicensure}{\textbf{Output:}}
\renewcommand{\algorithmicfunction}{\textbf{function}}

\definecolor{pipelinestepink}{HTML}{354450}
\tikzset{pipeline step/.style={circle,draw=pipelinestepink,text=pipelinestepink,
    line width=.65pt,fill=white,minimum size=.46cm,inner sep=0pt,
    font=\rmfamily\fontsize{9}{10.5}\selectfont}}
\DeclareRobustCommand{\pipelinestep}[1]{%
    \tikz[baseline=(step.base)]{\node[pipeline step,minimum size=.32cm,
        line width=.5pt,font=\rmfamily\fontsize{7.5}{9}\selectfont]
        (step) {#1};}}

\usepackage{bm}

\usepackage{scalerel}

\newcommand{\Hsquare}{%
  \text{\kern2\scriptspace\fboxsep=-.2pt\fbox{\rule{0pt}{1ex}\rule{1ex}{0pt}}\kern2\scriptspace}%
}

\def\epsilon{\ensuremath{\varepsilon}}

\def\presuper#1#2%
  {\mathop{}%
   \mathopen{\vphantom{#2}}^{#1}%
   \kern-\scriptspace%
   #2}
\def\surround#1#2#3%
  {\mathop{}%
   \mathopen{\vphantom{#2}}^{#1}%
   \kern-3\scriptspace%
   #2%
   \kern4\scriptspace%
   \mathopen{\vphantom{#2}}^{#3}%
   }

\usepackage{stmaryrd}

\newcommand{\oodbench}{OOD\xspace}
\newcommand{\mathbench}{MATH500\xspace}
\newcommand{\summeval}{SummEval\xspace}
\newcommand{\ifbench}{IF-Bench\xspace}
\newcommand{\gpqabench}{GPQA Diamond\xspace}
\newcommand{\ift}{IFT\xspace}
\newcommand{\opd}{OPD\xspace}
\newcommand{\iftmerge}{+\textsc{Merge}\xspace}
\newcommand{\iftopd}{+\textsc{OPD}\xspace}
\newcommand{\iftforce}{+\textsc{Force}\xspace}
\newcommand{\iftabort}{+\textsc{Abort}\xspace}

\newcommand{\openthinker}{OpenThinker\xspace}
\newcommand{\apriel}{Apriel\xspace}
\newcommand{\olmo}{Olmo3\xspace}
\newcommand{\qwen}{Qwen R1-D\xspace}
\newcommand{\gemini}{Gemini 3.1 Pro Preview\xspace}

\newcounter{algoline}[algorithm]

\usepackage[capitalize]{cleveref}

\crefformat{section}{\S#2#1#3}

\crefrangeformat{section}{\S#3#1#4\crefrangeconjunction\S#5#2#6}

\crefmultiformat{section}{\S#2#1#3}{\crefpairconjunction\S#2#1#3}{\crefmiddleconjunction\S#2#1#3}{\creflastconjunction\S#2#1#3}

\newcommand{\crefrangeconjunction}{--}

\crefname{listing}{Lst.}{listings}
\crefname{line}{Line}{Lines}
\crefname{appendix}{App.}{App.}

\newcommand{\app}[1]{%
	\ifbool{includeappendix}{\cref{#1}}{the appendix}%
}
\newcommand{\App}[1]{%
	\ifbool{includeappendix}{\cref{#1}}{The appendix}%
}

\title{Leveraging Instruction Tuning and Merging for Reasoning Model Adaptation}

\author{
Yu-Du Feng\thanks{Equal contribution.} \quad
Niels Mündler-Sasahara\footnotemark[1] \quad
\textbf{Mark Vero \quad Martin Vechev} \\
Department of Computer Science, ETH Zurich \\
\texttt{yufeng1@student.ethz.ch} \quad
\texttt{niels.muendler@inf.ethz.ch}\\
\raisebox{-0.2\height}{\includegraphics[height=1em]{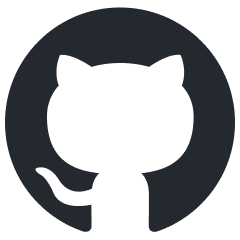}} \url{https://github.com/eth-sri/rlm-training-merging}
}

\iclrfinalcopy

\begin{document}

\maketitle
\lhead{}

\begin{abstract}
Reasoning language models (RLMs) demonstrate impressive performance by leveraging test-time compute in the form of reasoning tokens. However, this behavior makes adapting RLMs to new domains challenging and expensive. The reason is that further training can disturb the learned behavior and degrade model performance. This makes it difficult to leverage supervised fine-tuning data with human-written solutions: although it contains high-quality annotations, it lacks reasoning tokens.
In this work, we show how, despite this challenge, such data can be used efficiently for RLM adaptation. For this, we first use standard instruction tuning. Next, we leverage model merging to combine the instruction-tuned model with the original RLM, picking the merging ratio such that the resulting model's reasoning behavior on the target domain is recovered. We evaluate our method across four RLMs on coding and text summarization tasks, where it improves target-task performance by up to $11.0\%$ while preserving reasoning behavior and limiting the out-of-distribution score degradation to on average $0.7\%$. Importantly, our adaptations are efficient and economical, costing less than USD $\$10$ per model.
\end{abstract}

\section{Introduction}
\label{sec:introduction}

Reasoning language models (RLMs) have defined the frontier of language model capabilities by demonstrating impressive results on tasks such as mathematics and programming \citep{openaio1,qwen3}. RLMs are trained to produce reasoning traces that explore potential solutions or perform logical reasoning before producing a final answer \citep{grpo}. The dominant training recipes for RLMs are based on reinforcement learning with verifiable rewards (RLVR) \citep{deepseekr1,olmo3}. 
While this approach is unquestionably powerful, it is highly expensive: Training requires repeated model rollouts for each training step, with long rollouts requiring thousands of forward passes of the model \citep{ICLR2026_4badb55b}.

For non-reasoning models, adaptation to new domains and knowledge is typically achieved through instruction fine-tuning (\ift{}): The model is trained directly on task descriptions paired with high-quality task solutions, collected through scraping or human annotation \citep{zhang2025instructiontuninglargelanguage,lambert2024tulu3,sven}.
Large amounts of training data are available for IFT, which makes the approach cheap, effective, and widely applicable \citep{wei2022finetunedlanguagemodelszeroshot,chung2022scalinginstructionfinetunedlanguagemodels}.

However, the task solutions in IFT training data, unlike RLM outputs, lack reasoning traces, which creates a distributional mismatch when training RLMs using IFT. This mismatch can severely degrade general RLM performance \citep{lobo2025impact}.

\paragraph{This work: Leveraging \ift{} for training RLMs}
In this work, we show that this mismatch can be mitigated effectively with a two-step procedure. First, we perform standard \ift{} on the input-output pairs without reasoning traces. Second, we merge the resulting \ift{} checkpoint with the original reasoning model. The merge ratio is selected as the largest coefficient that preserves target-task reasoning on a small calibration set. In contrast to prior proposed methods, this procedure does not require a reliable verifier or data beyond the task-specific training set and is highly cost-effective.

We evaluate our approach on four open RLMs spanning different model families, sizes, and training paradigms: \openthinker{} 7B \citep{openthinker}, \qwen{} 7B \citep{deepseekr1}, \olmo{} 7B \citep{olmo3}, and \apriel{} 15B \citep{apriel}. We evaluate our method on Rust coding and text summarization and measure out-of-distribution performance on three unrelated benchmarks. On average, our method results in out-of-distribution performance loss of $0.7\%$, while resulting in target-task gains between $1.0\%$ and $11.0\%$. End-to-end adaptation costs between USD $\$1.07$ and $\$9.17$, reducing cost by $63\%$ to $99\%$ relative to our evaluated RL baselines.

\paragraph{Key contributions} Our key contributions are: (i) We identify and study a practical adaptation setting for RLMs where only input-output supervision is available, without verified reasoning traces, (ii) we propose a lightweight \ift{}-and-merge method that adapts an RLM while selecting the merge ratio using only reasoning behavior on target-task calibration data, and (iii) in our extensive evaluation, we show that our method effectively balances target-task performance with reasoning and out-of-distribution behavior.

\section{Background}\label{sec:background}
\vspace{-2mm}

In this section, we introduce (reasoning) language models, instruction fine-tuning, and model merging.

\paragraph{Language models}
An autoregressive language model (LM) is a statistical model of language with parameters $\theta$.
We can sample from a model $M_\theta$ given context $\rvc$ to obtain an \emph{answer} $\hat{\rvy} \sim{} M_\theta(\rvc)$.
Pre-trained large language models (LLMs) are LMs with billions of parameters, trained on trillions of tokens of training data \citep{gpt4,deepseekr1,qwen3}.
Through such pre-training, the models acquire a variety of general skills, in particular language understanding.

\begin{wrapfigure}[18]{r}{.48\textwidth}
\centering
\vspace{-7mm}
\begingroup
\input{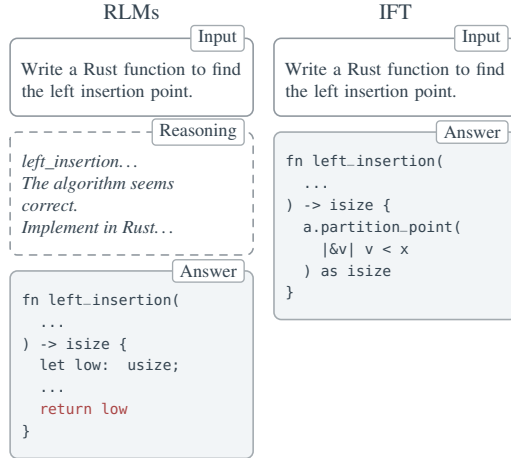}
\resizebox{\linewidth}{!}{%
\begin{tikzpicture}[
  x=1cm,y=1cm,text=pipeink,line cap=round,line join=round,
  font=\rmfamily\fontsize{7.5}{9}\selectfont,
  heading/.style={font=\rmfamily\fontsize{9}{10.5}\selectfont},
  snippet/.style={anchor=north west,align=left,inner sep=0pt,
    font=\ttfamily\fontsize{7.5}{9}\selectfont},
  code/.style={draw=pipeink!45,fill=pipecode,rounded corners=3pt,line width=.65pt}
]
\path[use as bounding box] (0,-5.28) rectangle (7.3,1.44);
\foreach \x/\titletext in {0/RLMs,3.8/IFT} {
  \node[heading] at (\x+1.75,1.25) {\titletext};
  \draw[input card]
    (\x,.87) rectangle (\x+3.5,-.25);
  \node[anchor=north west,align=left,text width=3.16cm,inner sep=0pt]
    at (\x+.16,.53)
    {Write a Rust function to find the left insertion point.};
  \node[output badge] at (\x+3.38,.87) {Input};
}
\draw[draw=pipeink!55,dashed,rounded corners=3pt,line width=.65pt]
  (0,-.48) rectangle (3.5,-2.25);
\node[output badge] at (3.38,-.48) {Reasoning};
\node[anchor=north west,align=left,inner sep=0pt,
  font=\rmfamily\itshape\fontsize{7.5}{9}\selectfont] at (.16,-.80)
  {left\_insertion\ldots\\The algorithm seems\\correct.\\Implement in Rust\ldots};
\draw[code] (0,-2.48) rectangle (3.5,-5.18);
\node[output badge] at (3.38,-2.48) {Answer};
\node[snippet] at (.16,-2.80) {fn left\_insertion(\\
  \ \ ...\\
  ) -> isize \{\\
  \ \ let low: usize;\\
  \ \ ...\\
  \ \ \textcolor{pipered}{return low}\\
  \}};
\draw[code] (3.8,-.48) rectangle (7.3,-3.18);
\node[output badge] at (7.18,-.48) {Answer};
\node[snippet] at (3.96,-.80) {fn left\_insertion(\\
  \ \ ...\\
  ) -> isize \{\\
  \ \ a.partition\_point(\\
  \ \ \ \ |\&v| v < x\\
  \ \ ) as isize\\
  \}};
\end{tikzpicture}%
}
\endgroup
\vspace{-5mm}
\caption{RLMs reason before answering; \ift{} targets start directly with the answer.}
\label{fig:reasoning-vs-ift}
\end{wrapfigure}

\paragraph{RLMs}
Reasoning language models (RLMs) are LLMs trained to generate a \emph{reasoning trace} $\mathbf{r}$ before producing the answer $\hat{\rvy}$. The reasoning trace can serve as internal scratch-pad, not scored for the final answer, and is separated from the answer by an end-of-reasoning token. 
RLMs are either trained via reinforcement learning with verifiable rewards (RLVR) \citep{openaio1,deepseekr1,qwen3} or through distillation of reasoning traces from a stronger model \citep{openthinker,deepseekr1}.
Empirically, this results in strong capabilities for logic-based tasks such as mathematics and coding \citep{grpo,deepseekr1}.

\paragraph{Supervised and instruction fine-tuning}

In practice, models can be adapted to specialized tasks by training them on novel knowledge \citep{farn2025safeguardfinetunedllmspre,lambert2024tulu3}.
Such adaptation is commonly achieved through supervised fine-tuning (SFT) \citep{stiennon2022learningsummarizehumanfeedback,sven}. SFT updates a model's parameters using a labeled dataset $\gD = \{(\rvx, \rvy)\}$ of input-output pairs, and trains the model to predict $\rvy$ for input $\rvx$. Instruction fine-tuning (\ift{}) is the special case of SFT where $\rvx$ is an instruction or task description and $\rvy$ is an answer or solution to the task \citep{muennighoff2024octopackinstructiontuningcode,chung2022scalinginstructionfinetunedlanguagemodels}. Large datasets of high-quality input-output pairs for a variety of target domains exist \citep{lambert2024tulu3,starcoder2} or can be synthesized \citep{unnatural,selfinstruct}, which allows model users to adapt models to a desired domain.



\paragraph{Model merging}
Model merging combines two sets of model weights, $\theta_1$ and $\theta_2$, into weights $\theta_\alpha$ using linear interpolation with a \emph{merging ratio} $\alpha$.
\[
    \theta_\alpha = \theta_1 + \alpha\Delta,\quad \alpha\in[0,1],\quad \Delta = \theta_2-\theta_1.
\]
Model merging can be used to combine model capabilities \citep{merge-transfer,ties} and recover forgotten behavior \citep{catastrophicforgetting,farn2025safeguardfinetunedllmspre}. Prior work shows that merging can produce capable models along various merging ratios of related low-loss training checkpoints \citep{ilharco2022editing,frankle2020lmc,catastrophicforgetting,yang2024modelmerging}. Non-linear merging variants exist as well \citep{ties,slerp}.

\begin{figure*}[t]
\centering
\begingroup
\input{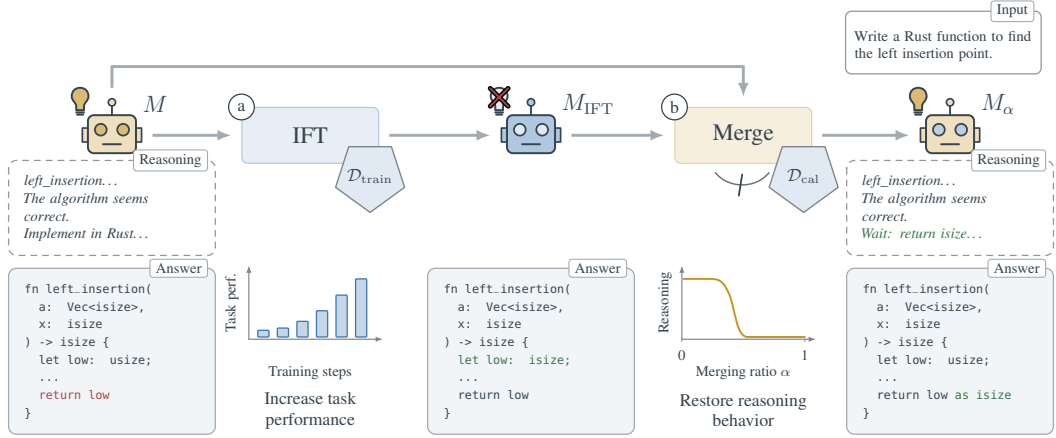}
\resizebox{\linewidth}{!}{%
\begin{tikzpicture}[
  x=1cm,y=1cm,
  font=\rmfamily\fontsize{9}{10.5}\selectfont,
  text=pipeink,
  line cap=round,line join=round,
  flow/.style={draw=pipeink!45,line width=1.5pt,-{Latex[length=2.5mm,width=2mm]}},
  connector/.style={flow,line cap=butt,shorten <=1.8mm,shorten >=1.8mm},
  process/.style={rounded corners=3pt,draw=pipeblue!45,fill=pipeblue!12,
    minimum width=2.45cm,minimum height=1cm,font=\rmfamily\fontsize{12}{14}\selectfont},
  data/.style={regular polygon,regular polygon sides=5,shape border rotate=180,
    draw=pipeink!65,line width=.7pt,fill=pipeblue!15,minimum size=1.30cm,
    inner sep=1pt,font=\rmfamily\fontsize{9}{10.5}\selectfont},
  trace/.style={draw=pipeink!55,dashed,rounded corners=4pt,line width=.65pt},
  code/.style={draw=pipeink!45,fill=pipecode,rounded corners=4pt,line width=.65pt},
  prose/.style={anchor=north west,align=left,inner sep=0pt,text width=3.16cm,
    font=\rmfamily\itshape\fontsize{7.5}{9}\selectfont},
  snippet/.style={anchor=north west,align=left,inner sep=0pt,
    font=\ttfamily\fontsize{7.5}{9}\selectfont}
]
\path[use as bounding box] (-.55,-3.17) rectangle (18.35,4.77);
\pic (base) at (1.4,2.3) {robot={pipegold!30/pipegold!65}};
\pic (tuned) at (8.9,2.3) {robot={pipeblue!40/pipeblue!15}};
\pic (merged) at (16.4,2.3) {robot={pipegold!30/pipeblue!35}};
\node[anchor=west,font=\fontsize{12}{14}\selectfont] at (1.83,2.86) {$M$};
\node[anchor=west,font=\fontsize{12}{14}\selectfont] at (9.33,2.86) {$\mift$};
\node[anchor=west,font=\fontsize{12}{14}\selectfont] at (16.83,2.86) {$\malpha$};
\pic at (.83,2.82) {bulb={pipegold!65}};
\pic at (8.33,2.82) {bulb={pipeink!12}};
\draw[black,line width=2.5pt] (8.17,2.82) -- (8.49,3.14)
  (8.17,3.14) -- (8.49,2.82);
\draw[pipered,line width=1pt] (8.17,2.82) -- (8.49,3.14)
  (8.17,3.14) -- (8.49,2.82);
\pic at (15.83,2.82) {bulb={pipegold!65}};

\node[process] (ift) at (4.95,2.30) {IFT};
\node[process,fill=pipegold!15,draw=pipegold!50] (merge) at (12.7,2.30) {Merge};
\foreach \stage/\letter in {ift/a,merge/b} {
  \node[pipeline step]
    at (\stage.north west) {\letter};
}
\draw[connector] (base-east) -- (ift.west);
\draw[connector] (ift.east) -- (tuned-west);
\draw[connector] (tuned-east) -- (merge.west);
\draw[connector] (merge.east) -- (merged-west);
\draw[connector] (base-top) -- (1.4,3.53) -- (12.7,3.53) -- (merge.north);
\node[data] at (6.02,1.57) {};
\node[data] at (13.78,1.57) {};
\node[font=\fontsize{9}{10.5}\selectfont] at (6.02,1.57) {$\dtrain$};
\node[font=\fontsize{9}{10.5}\selectfont] at (13.78,1.57) {$\dcal$};
\draw[pipeink!75,line width=.75pt] (12.20,1.63)
  .. controls (12.44,1.34) and (12.92,1.34) .. (13.16,1.63);
\draw[pipeink,line width=.9pt] (12.68,1.60) -- (12.62,1.22);

\draw[input card] (14.53,3.40) rectangle (18.03,4.52);
\node[anchor=north west,align=left,text width=3.16cm,inner sep=0pt,
  font=\rmfamily\fontsize{7.5}{9}\selectfont]
  at (14.69,4.18) {Write a Rust function to find the left insertion point.};
\node[output badge] at (17.91,4.52) {Input};

\foreach \x in {0,7.5,15} {
  \draw[code] (\x-.45,-.08) rectangle +(3.70,-2.97);
  \node[output badge] at (\x+3.12,-.08) {Answer};
}
\foreach \x in {0,15} {
  \draw[trace] (\x-.45,1.85) rectangle +(3.70,-1.70);
  \node[output badge] at (\x+3.12,1.85) {Reasoning};
}
\node[prose] at (-.18,1.59) {left\_insertion\ldots\\
  The algorithm seems\\correct.\\Implement in Rust\ldots};
\node[prose] at (14.82,1.59) {left\_insertion\ldots\\
  The algorithm seems\\correct.\\
  \textcolor{pipegreen}{Wait: return isize\ldots}};

\node[snippet] at (-.17,-.36) {fn left\_insertion(\\
  \ \ a: Vec<isize>,\\
  \ \ x: isize\\
  ) -> isize \{\\
  \ \ let low: usize;\\
  \ \ \ldots\\
  \ \ \textcolor{pipered}{return low}\\
  \}};
\node[snippet] at (7.33,-.36) {fn left\_insertion(\\
  \ \ a: Vec<isize>,\\
  \ \ x: isize\\
  ) -> isize \{\\
  \ \ \textcolor{pipegreen}{let low: isize;}\\
  \ \ \ldots\\
  \ \ return low\\
  \}};
\node[snippet] at (14.83,-.36) {fn left\_insertion(\\
  \ \ a: Vec<isize>,\\
  \ \ x: isize\\
  ) -> isize \{\\
  \ \ let low: usize;\\
  \ \ \ldots\\
  \ \ return low \textcolor{pipegreen}{as isize}\\
  \}};

\begin{scope}[shift={(3.85,-1.35)}]
  \draw[pipeink!65,line width=.6pt,-{Latex[length=1.3mm]}] (0,0) -- (2.40,0);
  \draw[pipeink!65,line width=.6pt,-{Latex[length=1.3mm]}] (0,0) -- (0,1.32);
  \foreach \x/\h in {.16/.12,.51/.16,.86/.28,1.21/.47,1.56/.75,1.91/1.04}
    \draw[fill=pipeblue!30,draw=pipeblue,line width=.65pt,rounded corners=.5pt]
      (\x,.04) rectangle +(0.20,\h);
  \node[font=\rmfamily\fontsize{7.5}{9}\selectfont,anchor=north] at (1.10,-.36) {Training steps};
  \node[font=\rmfamily\fontsize{7.5}{9}\selectfont,rotate=90] at (-.30,.66) {Task perf.};
\end{scope}
\node[align=center,font=\rmfamily\fontsize{9}{10.5}\selectfont]
  at (4.95,-2.62) {Increase task\\performance};

\begin{scope}[shift={(11.60,-1.35)}]
  \draw[pipeink!65,line width=.6pt,-{Latex[length=1.3mm]}] (0,0) -- (2.40,0);
  \draw[pipeink!65,line width=.6pt,-{Latex[length=1.3mm]}] (0,0) -- (0,1.32);
  \draw[pipegold,line width=1pt]
    (.03,1.08) -- (.50,1.08)
    .. controls (.70,1.08) and (.75,1.02) .. (.85,.80)
    .. controls (.96,.53) and (.97,.04) .. (1.18,.04)
    -- (2.20,.04);
  \foreach \x/\ticktext in {0/0,2.20/1} {
    \draw[pipeink!65,line width=.5pt] (\x,0) -- (\x,-.04);
    \node[anchor=north,font=\rmfamily\fontsize{7.5}{9}\selectfont]
      at (\x,-.07) {\ticktext};
  }
  \node[font=\rmfamily\fontsize{7.5}{9}\selectfont,rotate=90]
    at (-.30,.66) {Reasoning};
  \node[anchor=north,font=\rmfamily\fontsize{7.5}{9}\selectfont]
    at (1.10,-.36) {Merging ratio $\alpha$};
\end{scope}
\node[align=center,font=\rmfamily\fontsize{9}{10.5}\selectfont]
  at (12.70,-2.62) {Restore reasoning\\behavior};
\end{tikzpicture}%
}
\endgroup
    \vspace{-4mm}
\caption{We propose a lightweight two-step pipeline for RLM adaptation to a target task, here Rust coding. It consists of \pipelinestep{a} training on IFT data and then \pipelinestep{b} merging the resulting model $\mift$ with the original model $M$, picking the maximal merging ratio that preserves reasoning behavior. It balances capabilities obtained through \ift{} training with observed reasoning behavior.}
    \label{fig:overview}
    \label{fig:mergeoverview}
\label{fig:ift-merge-pipeline}
    \vspace{-2mm}
\end{figure*}

\section{Training RLMs via \ift{} and Model Merging}
\label{sec:method}

In this section, we outline the key challenges and describe our core pipeline for adapting RLMs. The key insight behind our method is that traditional \ift{} can still be used for RLMs when carefully combined with a model-merging technique that preserves the model's reasoning behavior.

\paragraph{Main challenges}
The main challenge for adapting RLMs lies in the learned reasoning behavior.
Training the model without reasoning traces as in \ift{} creates a conflict with the learned output distribution: the model is trained to exhibit reasoning during rollouts, but such reasoning traces are absent from the training data available during \ift{} as shown in \cref{fig:reasoning-vs-ift}.
This poses a challenge because training RLMs without reasoning traces can lead to performance degradation and catastrophic forgetting \citep{twist2026reasoningtracecollapseevaluatingloss}. Meanwhile, continuing RLVR on the target domain is expensive \citep{relift,grpo}, and works predominantly in domains where reliable verifiers are available \citep{cotcotchainofthoughthelps,rewardhacking}.
Our work addresses these challenges by training RLMs on IFT data without requiring reasoning traces. As such, our method can leverage widely available \ift{} datasets to teach models novel knowledge while preserving learned reasoning behavior.

\paragraph{Motivation}
Our key insight is twofold: First, IFT is a proven and tested method for adapting standard LMs \citep{muennighoff2024octopackinstructiontuningcode, sven,stiennon2022learningsummarizehumanfeedback}. Second, reasoning behavior 
is learned late in the model training process \citep{olmo3,qwen3}. Its absence correlates in general with capability degradation \citep{lobo2025impact} and might serve as a canary for general model forgetting of learned skills. Model merging is a lightweight intervention that mitigates model forgetting \citep{catastrophicforgetting} and at the same time can recover reasoning behavior when adequately scaled \citep{zhang2024unconstrained}.

\paragraph{Overview}
Our core method for RLM training consists of a lightweight two-step pipeline, visualized in \cref{fig:overview}: We first perform standard \ift{} on RLM $M$ using dataset $\dtrain$, which produces a fine-tuned model $\mift$ with higher task accuracy but potentially compromised reasoning behavior and other forgotten capabilities. In the shown example, \apriel{} correctly inserts missing type casts in Rust code after fine-tuning, but does not emit any reasoning about the code. We then merge $M$ and $\mift$, picking the maximal merge ratio $\alpha$ that restores model reasoning on calibration dataset $\dcal$. The merge produces $\malpha$, a reasoning model with maintained or improved task capability. In the example, the merged \apriel{} then reasons about the typing mismatch and required cast and correctly inserts it in the resulting code. Complete outputs are presented in \cref{app:rust-case-study-outputs}.
By design, our method requires no reliable verifiers, expensive model rollouts, or stronger RLMs.

\paragraph{Datasets}
We use $\dtrain$, $\dcal$, and $\dtest$ for fine-tuning, merge-ratio selection, and final evaluation, respectively. Each task example is an input-output pair $(i,o)$, where $i$ is the task input and $o$ is the target output. For a reasoning model, we write a sampled response as $(r,\hat{o}) \sim M(i)$, where $r$ is the reasoning trace and $\hat{o}$ is the final answer.

\paragraph{Fine-tuning}
We first fine-tune the RLM $M$ on $\dtrain$ using standard \ift{} \citep{ift}.
Since our dataset $\dtrain$ contains only $(i,o)$ pairs and does not contain the reasoning trace $r$, we serialize each training target as $(\varepsilon,o)$: the empty reasoning trace $\varepsilon$ followed by the target output. As a result, our pipeline obtains model $\mift$ with weights $\theta_\text{\ift{}}$.

Because the fine-tuning targets contain an empty reasoning trace, this step can reduce the model's reasoning behavior: $\mift$ may directly output the final answer on examples where the original reasoning model would have produced reasoning tokens.

\paragraph{Merging}
To recover reasoning behavior, we perform a model merge between the original reasoning model $M$ and $\mift$.
We select the merging ratio $\alpha$ by searching over model reasoning on a target-task calibration set $\dcal$. Since $M$ is an RLM in our setting, it has a full calibration reasoning rate, i.e., $\rho(M,\dcal)=1$. For any model $M'$ and dataset $\gD$, we define the reasoning rate $\rho(M', \gD)$ as the fraction of examples for which the model produces a non-empty reasoning trace.
\begin{equation}
    \rho(M',\gD)
    =
    \frac{1}{|\gD|}
    \sum_{(i,o)\in \gD}
    \mathbf{1}\{r_i \neq \varepsilon\},
    \qquad
    (r_i,\hat{o}_i) \sim M'(i).
\end{equation}

\begin{wrapfigure}{r}{0.43\linewidth}
    \vspace{-9mm}
\begin{minipage}{\linewidth}
\begin{algorithm}[H]
\caption{Our Adaptation Algorithm}
\label{alg:ift-merge}
\footnotesize
\begin{algorithmic}[1]
\REQUIRE RLM $M$; $\dtrain, \dcal$; $\rho_\text{min}$; $K$
\ENSURE Adapted RLM $\malpha[\alpha^\star]$
\item[] \vspace{1mm}\hspace{-5mm} Step 1: Standard \ift{} \vspace{0.5mm}
\STATE $\mift \leftarrow \textsc{\ift{}}(M, \dtrain)$
\item[] \vspace{1mm}\hspace{-5mm} Step 2: Merging \vspace{0.5mm}
\STATE $\mathcal{A}_K \leftarrow \{j/K \mid j \in \{0,\ldots,K\} \}$
\STATE $\alpha^\star \leftarrow 0$
\FOR{$\alpha \in \mathcal{A}_K$}
    \STATE $\malpha \leftarrow \textsc{Merge}(M, \mift, \alpha)$
    \STATE $\rho_\alpha \leftarrow \rho(\malpha, \dcal)$
    \IF{$\rho_\alpha \geq \rho_\text{min}$}
        \STATE $\alpha^\star \leftarrow \alpha$
    \ENDIF
\ENDFOR
\STATE \textbf{return} $\malpha[\alpha^\star]$
\end{algorithmic}
\end{algorithm}
\end{minipage}
    \vspace{-4mm}
\end{wrapfigure}

We design a search procedure to select the model closest to the \ift{} checkpoint that preserves target-task performance gains while recovering the model's target-task reasoning. Let $\rho_\text{min}$ be the minimum acceptable reasoning rate; in our experiments, $\rho_\text{min}=1.0$. We are then looking for the largest merge ratio whose target-task calibration reasoning rate remains acceptable, i.e.,
\begin{equation}
    \alpha^\star
    =
    \max_{\alpha \in \mathcal{A}_K}
    \left\{
        \alpha
        :
        \rho_\text{min} \leq \rho(\malpha,\dcal)
    \right\}.
\end{equation}
This objective favors the model closest to $\mift$ among those that still preserve enough reasoning behavior.

We describe the algorithm in \cref{alg:ift-merge}.
After \ift{}, we evaluate a small uniform grid of merge ratios. For a grid resolution $K$, we use $\mathcal{A}_K = \{0, 1/K, 2/K, \ldots, 1\}$. The point $\alpha=0$ is the original model $M$ and is guaranteed to satisfy $\rho(M,\dcal)=1$, while $\alpha=1$ is the standard \ift{} model \mift{}. We evaluate all points on $\dcal$, and select the largest $\alpha$ whose reasoning rate is at least $\rho_\text{min}$.

\paragraph{Optimizations}
We apply two optimizations to speed up this process: running calibration only on the first few tokens and using binary search in place of exhaustive grid search.

First, we observe that, when \mift{} does not reason, it emits the end-of-reasoning token immediately at the beginning of its output and does not fall back into reasoning afterwards. We leverage this to speed up the merge-ratio selection: for each calibration input, we generate only the first few tokens after the reasoning start token and check whether the model begins a non-empty reasoning trace. This short-prefix evaluation is enough to reliably determine $\rho(\malpha,\dcal)$ for the grid search.

Second, we reduce the computational effort for the search in \cref{alg:ift-merge}. We observe that the reasoning rate decreases monotonically as the merge ratio increases. This allows us to employ binary search instead of grid search in compute-constrained settings. We use binary search to identify the largest merge ratio at which reasoning on the calibration dataset is at least the minimum threshold.

\section{Experimental Evaluation}
\label{sec:experiments}

In this section, we evaluate the trade-off achieved by our method between target-task performance and general reasoning capabilities across four reasoning models and two datasets.

\subsection{Experimental Setup}


\paragraph{Models}
We compare four recent RLMs: \openthinker{} 7B \citep{openthinker}, \qwen{} 7B \citep{deepseekr1}, \olmo{} 7B \citep{olmo3}, and \apriel{} 15B \citep{apriel}. These models cover diverse training approaches and model families: \openthinker{} and \qwen{} are post-trained variants of Qwen2.5 7B \citep{qwen2.5}, obtaining reasoning capabilities via distillation on reasoning traces generated by DeepSeek R1 \citep{deepseekr1}. \apriel{} and \olmo{} are based on pre-trained base models and post-trained via \ift{}, DPO (only \olmo{}) and RLVR.

\paragraph{Training datasets}
We train all RLMs on two tasks: Rust coding and text summarization.
While Rust coding requires logical reasoning and knowledge of the Rust language, text summarization requires deep textual understanding. Additionally, text summarization represents a domain without reliable verifiers, for which standard RL may be subject to reward hacking \citep{rewardhacking}.

We curate two datasets as \dtrain{}.
(1) \emph{Rust coding}: A set of synthetic single-function coding tasks in the Rust language \citep{oxen-rust-grpo}, filtered to test-passing instances, resulting in 6,761 training samples. We reserve 750 samples for \dval{}.
(2) \emph{Text summarization}: The 10,000-sample Reddit TLDR split of \citet{stiennon2022learningsummarizehumanfeedback}, a collection of user-written Reddit posts with self-written summaries. We reserve 400 samples for \dval{}.
Additional details on dataset construction in \cref{app:rust-details}.

\paragraph{Metrics}
For Rust coding, we use the Rust-translated MBPP \citep{mbppdataset} and measure pass@1 \citep{humaneval}, reported as \emph{MBRP}.
For text summarization, we report the average of fluency, consistency, relevance, and coherence on the CNN/Daily Mail split of the \summeval dataset \citep{summeval} as \emph{\summeval}.
Precise score definitions are in \cref{app:scores-details}.
The training and test datasets are intentionally drawn from slightly different distributions while retaining similar task formats to measure generalization beyond training settings.

We further evaluate out-of-distribution model forgetting and general reasoning capabilities via independent benchmarks, denoted as \emph{\oodbench{}}. \oodbench is the equal-weight average of \mathbench \citep{math500}, \ifbench{} \citep{ifbench}, and \gpqabench \citep{gpqa}.
Separately, we report the rate of non-empty reasoning traces on the evaluation dataset (\emph{Reasoning}).

\paragraph{Methods}
Unless otherwise mentioned, our method (\emph{\iftmerge}) is implemented with linear merging, binary search, up to eight search steps, with $\rho_\text{min} = 1$.
We report the unadapted RLM (\emph{Unadapted}), instruction fine-tuning (\emph{\ift{}}), and two additional baselines:
First, On-Policy Distillation (\emph{\iftopd{}}) \citep{onpolicydistill,lu2025onpolicydistillation} uses an unrelated instruction fine-tuning dataset \citep{lambert2024tulu3} to distill unadapted RLM behavior into the student model for a small number of steps. Second, we report RL-based methods (\emph{RL}).
For Rust, we use the test suites of the training dataset as a verifiable reward for RLVR \citep{grpo}. For text summarization, we implement RLAIF \citep{rlaif}, using each unadapted model as an output judge for generated outputs.


\paragraph{Protocol}
For all methods, we tune their hyperparameters in a grid sweep against \dval{}.
We tune the learning rate, batch size, and number of epochs for each combination of model, method, and dataset, and report the hyperparameters in \cref{app:ift-hyperparameters}. 
Each setting uses two training seeds, with 10 samples for Rust, two for summarization, and one per \oodbench{} task.

For scores, we \textbf{boldface} the best method per model and \underline{underline} insignificantly different methods, determined via a one-sided paired Wilcoxon signed-rank permutation test with Holm correction ($p_\mathrm{adj}\geq0.05$). We \textbf{boldface} the lowest cost.
In all plots, we draw bands indicating the minimum and maximum average scores obtained by each training run.

\begin{table*}[t]
\definecolor{costgray}{gray}{0.625}
\centering
\caption{
Across four models, our method ({\setlength{\fboxsep}{1.5pt}\colorbox{mergehighlightblue}{\iftmerge{}}}) improves both target tasks in all eight settings and preserves mean Reasoning at $99.7\%$ and $99.7\%$ on Rust (MBRP) and text summarization (SE). Mean OOD changes are $+0.60\%$ and $-1.94\%$, respectively. Costs are lower than RL and OPD.}
\label{tab:main-score-cost}
\small
\setlength{\tabcolsep}{3pt}
\resizebox{\linewidth}{!}{%
\begin{NiceTabular}{l@{\hspace{.5cm}}lrrrrcrrrr}
\toprule
& & \multicolumn{4}{c}{Rust coding} & & \multicolumn{4}{c}{Text summarization} \\
\cmidrule(lr){3-6}\cmidrule(lr){8-11}
Model & Method & MBRP & \oodbench{} & Reasoning & Cost & \hspace{.5cm} & SE & \oodbench{} & Reasoning & Cost \\
\midrule
\multirow{5}{*}{\openthinker{} 7B}
 & Unadapted & $56.9$ & $45.4$ & $\mathbf{100.0\%}$ & \textcolor{costgray}{$\$0.00$} &  & $83.8$ & $\underline{46.9}$ & $\underline{99.7\%}$ & \textcolor{costgray}{$\$0.00$} \\
 & RL & $58.6$ & $\mathbf{47.3}$ & $\mathbf{100.0\%}$ & $\$22.63$ &  & $30.9$ & $\underline{47.6}$ & $\mathbf{100.0\%}$ & $\$50.61$ \\
 & IFT & $57.3$ & $32.9$ & $0.0\%$ & \textcolor{costgray}{$\$2.55$} &  & $\mathbf{91.6}$ & $20.0$ & $0.0\%$ & \textcolor{costgray}{$\$0.81$} \\
 & \quad + OPD & $55.6$ & $45.2$ & $\mathbf{100.0\%}$ & $\$3.32$ &  & $74.8$ & $\mathbf{47.8}$ & $11.8\%$ & $\$1.65$ \\
 & \quad \rowcolor{mergehighlightblue} \iftmerge{} & $\mathbf{60.7}$ & $46.0$ & $98.6\%$ & $\mathbf{\$2.84}$ &  & $88.2$ & $\underline{45.0}$ & $\mathbf{100.0\%}$ & $\mathbf{\$1.08}$ \\
\midrule
\multirow{5}{*}{\qwen{} 7B}
 & Unadapted & $30.6$ & $\underline{51.0}$ & $\mathbf{100.0\%}$ & \textcolor{costgray}{$\$0.00$} &  & $78.3$ & $\underline{51.9}$ & $\mathbf{100.0\%}$ & \textcolor{costgray}{$\$0.00$} \\
 & RL & $31.8$ & $\mathbf{51.3}$ & $\mathbf{100.0\%}$ & $\$18.85$ &  & $40.8$ & $\mathbf{52.3}$ & $\mathbf{100.0\%}$ & $\$10.56$ \\
 & IFT & $\mathbf{50.1}$ & $48.6$ & $63.5\%$ & \textcolor{costgray}{$\$3.85$} &  & $81.3$ & $46.0$ & $0.8\%$ & \textcolor{costgray}{$\$3.67$} \\
 & \quad + OPD & $44.1$ & $\underline{50.0}$ & $\mathbf{100.0\%}$ & $\$5.00$ &  & $\mathbf{83.6}$ & $\underline{51.8}$ & $\mathbf{100.0\%}$ & $\$4.38$ \\
 & \quad \rowcolor{mergehighlightblue} \iftmerge{} & $41.6$ & $\underline{50.8}$ & $\mathbf{100.0\%}$ & $\mathbf{\$4.09}$ &  & $79.3$ & $\underline{49.9}$ & $\underline{99.9\%}$ & $\mathbf{\$3.86}$ \\
\midrule
\multirow{5}{*}{\olmo{} 7B}
 & Unadapted & $46.2$ & $59.9$ & $\mathbf{100.0\%}$ & \textcolor{costgray}{$\$0.00$} &  & $88.9$ & $\mathbf{61.6}$ & $\mathbf{100.0\%}$ & \textcolor{costgray}{$\$0.00$} \\
 & RL & $46.3$ & $\mathbf{63.1}$ & $\mathbf{100.0\%}$ & $\$37.41$ &  & $88.8$ & $\underline{61.4}$ & $\mathbf{100.0\%}$ & $\$98.77$ \\
 & IFT & $\mathbf{50.0}$ & $59.4$ & $\mathbf{100.0\%}$ & \textcolor{costgray}{$\$1.74$} &  & $\mathbf{92.4}$ & $58.6$ & $\mathbf{100.0\%}$ & \textcolor{costgray}{$\$0.93$} \\
 & \quad + OPD & $45.8$ & $60.3$ & $\mathbf{100.0\%}$ & $\$2.82$ &  & $90.9$ & $\underline{60.4}$ & $\mathbf{100.0\%}$ & $\$1.18$ \\
 & \quad \rowcolor{mergehighlightblue} \iftmerge{} & $\mathbf{50.0}$ & $59.4$ & $\mathbf{100.0\%}$ & $\mathbf{\$1.88}$ &  & $\mathbf{92.4}$ & $58.6$ & $\mathbf{100.0\%}$ & $\mathbf{\$1.07}$ \\
\midrule
\multirow{5}{*}{\apriel{} 15B}
 & Unadapted & $62.8$ & $56.9$ & $\mathbf{100.0\%}$ & \textcolor{costgray}{$\$0.00$} &  & $95.6$ & $\underline{58.4}$ & $\mathbf{100.0\%}$ & \textcolor{costgray}{$\$0.00$} \\
 & RL & $64.9$ & $\underline{58.3}$ & $\mathbf{100.0\%}$ & $\$25.70$ &  & $81.8$ & $\mathbf{58.9}$ & $\mathbf{100.0\%}$ & $\$26.53$ \\
 & IFT & $63.6$ & $56.5$ & $33.1\%$ & \textcolor{costgray}{$\$8.77$} &  & $94.0$ & $43.4$ & $35.2\%$ & \textcolor{costgray}{$\$1.85$} \\
 & \quad + OPD & $63.4$ & $56.8$ & $\mathbf{100.0\%}$ & $\$11.34$ &  & $\mathbf{97.2}$ & $53.9$ & $\mathbf{100.0\%}$ & $\$3.60$ \\
 & \quad \rowcolor{mergehighlightblue} \iftmerge{} & $\mathbf{66.5}$ & $\mathbf{59.4}$ & $\mathbf{100.0\%}$ & $\mathbf{\$9.17}$ &  & $\underline{97.0}$ & $\underline{57.7}$ & $98.9\%$ & $\mathbf{\$2.25}$ \\
\bottomrule
\end{NiceTabular}
}
\vspace{1mm}
\vspace{-3mm}
\end{table*}
\subsection{Main Results}
\label{sec:experiments-main}
Overall, we find that our method increases target task performance, maintains \oodbench{} performance and reasoning behavior, and is significantly cheaper than comparable baselines.

\vspace{-1mm}
\paragraph{Target-task improvements}
Our method \iftmerge{} improves the target score in all eight model--task settings,
averaging $+5.56\%$ on Rust and $+2.56\%$ on text summarization.
It maintains the performance gains of \ift{}, on which it is based, with $-0.55\%$ and $-0.60\%$ on average, respectively, and performs better for 2 of 4 models on Rust, and better or tied for 2 models in text summarization. Meanwhile,
\iftopd{} reduces the Rust score relative to \ift{} for all four models.
In \cref{app:summeval-detailed}, we find that \ift{} in text summarization mostly changes consistency but degrades relevance, which is recovered by \iftmerge{} and \iftopd{}.
For Rust, RL improves target task performance consistently by $1.26\%$.
The average decrease of RL on text summarization is $-26.1\%$. For all models, text scores decline relative to the unadapted model, most significantly \openthinker{} and \qwen{}. We link this to the poor supervision signal in RLAIF: Both affected models often failed to prefer human-written summaries to their own or even to produce parseable grading scores during rollouts.

\vspace{-2mm}
\paragraph{Preservation of \oodbench{} and Reasoning}
Direct \ift{} lowers \oodbench{} performance in all eight settings by an average of $3.94\%$ on Rust and $12.72\%$ on text summarization. Meanwhile, mean
reasoning rates drop to $49.2\%$ and $34.0\%$. Merging recovers the unadapted models' mean OOD scores, inducing changes of only $+0.60\%$ and $-1.94\%$ on average.
At the same time, our method restores reasoning rates to $99.7\%$ on both tasks.
The majority of \oodbench{} losses occur from reasoning-heavy \mathbench{}, we provide additional details in \cref{app:ood-detailed}.
RL preserves reasoning completely and obtains the highest mean OOD scores. This comes at the price of significantly higher cost.

\vspace{-2mm}
\paragraph{Runtime and cost}
RL is the most expensive evaluated method in all eight settings. Even though we apply only 100 training steps, whereas typical setups use several thousand \citep{deepseekr1,olmo3}, it averages \SI{7.9}{\hour} for Rust coding. For RLAIF in text summarization, the average is \SI{14.1}{\hour}, as additional model inference is required for model scoring. At a representative rental price of USD $\$3.31$ per H200 GPU per hour \citep{silicondatah200}, this results in an average RL training costs of USD $\$26.15$ and $\$46.62$ for Rust coding and text summarization, respectively.

\ift{} takes \SI{76.6}{\minute} for Rust coding and \SI{32.9}{\minute} for text summarization on average.
Because our method \iftmerge{} employs only lightweight operations, it takes only \SI{4.9}{\minute} on average for Rust and \SI{4.5}{\minute} for text summarization on top of \ift{} time.
In cost, this adds an average of only USD $\$0.27$ and $\$0.25$ to the corresponding \ift{} costs of USD $\$4.23$ and $\$1.82$, respectively.

In contrast, \iftopd{} requires additional training steps, adding USD $\$1.39$ and $\$0.89$ for Rust coding and text summarization, respectively, $5.2$ and $3.6$ times more than our merging. In practice, this cost is paid several times, as the setup requires an additional hyperparameter search for learning rate and number of steps to conduct for OPD. This search is included in the costs of our merge ratio search, but not in the OPD costs.

\paragraph{Other methods}
Prior works have suggested training RLMs using self-distillation (SDFT) \citep{sdft} or reverse loss on reasoning traces from golden answers (VeriFree) \citep{verifree}.
They require substantially more resources than our approach. For example, SDFT reports an expected cost of 4 times the cost of \ift{}. In our experiments, it used over 20 times our GPU resources: \SI{4.5}{\hour} on 4 H200 GPUs. We estimate VeriFree would require around \SI{70}{\hour} on 4 H200 GPUs.
We report our attempts to reproduce both methods in \cref{appsec:more-baseline}.

\subsection{Ablation and Additional Analysis}
\label{sec:experiments-ablation}

\paragraph{Merging ratio choice}
Our method is based purely on reasoning behavior, yet it preserves \oodbench{} performance. We investigate the relationship of reasoning behavior with model performance across a five-ratio uniform grid. As shown in the left panels of \cref{fig:merge-ratio-target-score}, across the 40 available points, a 10-point decrease in reasoning rate is associated with a $0.94$-point decrease in \oodbench{} performance within model--task trajectories. Further, reasoning rate typically degrades quickly after a model-specific threshold. Even though we choose a maximal merge ratio that preserves reasoning, the ratio selected by our method retains reasoning rates between $98.6\%$ and $100\%$.

Meanwhile, selected ratios fall within one percent point of the best evaluated target score in 5 of 8 model--task settings, with a median gap of $0.60\%$. Overall, we find that our method selects high-reasoning points that provide a reliable trade off between generalization and target task learning.
We analyse the change in reasoning length in \cref{app:reasoning-length} and find no clear relationship with model performance, in contrast to reasoning behavior.

\begin{figure*}[t]
    \centering
    \includegraphics[width=1\linewidth]{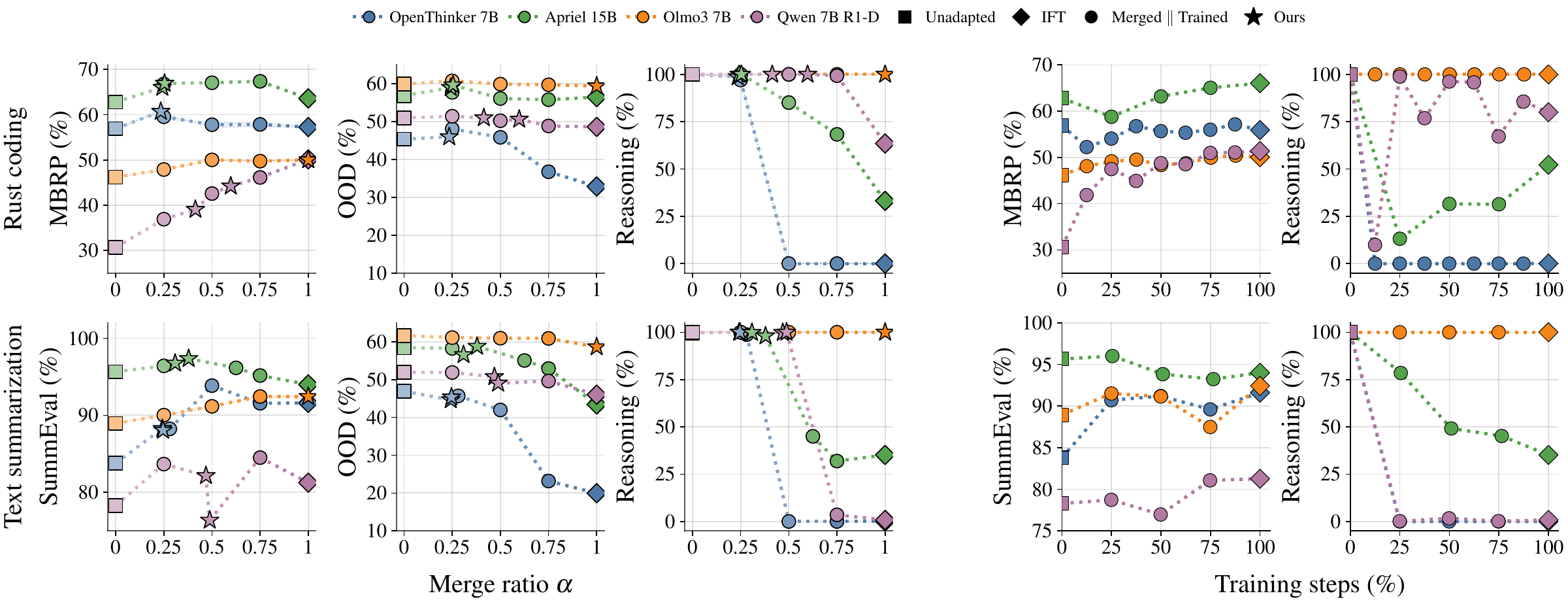}
    \caption{(Left) Across merge ratios, a 10-point reasoning decrease is associated with a 0.94-point \oodbench{} decrease. Ratios selected by \iftmerge{} retain $98.6$ to $100\%$ reasoning. (Right) During \ift{} training, target-task performance increases slowly, while reasoning behavior quickly collapses.}
    \label{fig:rust-grid-reasoning}
    \label{fig:merge-ratio-target-score}
    \label{fig:training-trajectories}
\vspace{-2mm}
\end{figure*}

\begin{table*}[t]
\centering
\caption{Across the models that exhibited reasoning loss, aborting before reasoning loss (\iftabort{}) falls short of the \ift{} target score by $7.14\%$ on Rust and $11.32\%$ on text summarization. Forcing \mift{} to emit a non-empty trace (\iftforce{}) lowers \oodbench{} performance relative to the unadapted model by $1.05\%$ and $9.89\%$, respectively.}
\label{tab:control-score-comparison}
\small
\setlength{\tabcolsep}{2pt}
\resizebox{.8\linewidth}{!}{%
\begin{NiceTabular}{l@{\hspace{.15cm}}cccc@{\hspace{.15cm}}cccc@{\hspace{.15cm}}cccc}
\toprule
& \multicolumn{4}{c}{OpenThinker} & \multicolumn{4}{c}{Apriel} & \multicolumn{4}{c}{Qwen R1-D} \\
\cmidrule(lr){2-5}\cmidrule(lr){6-9}\cmidrule(lr){10-13}
Method & Rust & OOD & Text & OOD & Rust & OOD & Text & OOD & Rust & OOD & Text & OOD \\
\midrule
Unadapted & $56.9$ & $45.4$ & $83.8$ & $\mathbf{46.9}$ & $62.8$ & $56.9$ & $95.6$ & $\mathbf{58.4}$ & $30.6$ & $\underline{51.0}$ & $78.3$ & $\mathbf{51.9}$ \\
IFT & $57.3$ & $32.9$ & $\mathbf{91.6}$ & $20.0$ & $63.6$ & $56.5$ & $94.0$ & $43.4$ & $\mathbf{50.1}$ & $48.6$ & $\mathbf{81.3}$ & $46.0$ \\
\hspace{5mm} + Abort & $57.4$ & $\mathbf{47.8}$ & $83.4$ & $\underline{46.9}$ & $63.2$ & $\underline{58.9}$ & $72.2$ & $\underline{56.6}$ & $28.9$ & $\mathbf{52.2}$ & $77.3$ & $\underline{51.3}$ \\
\hspace{5mm} + Force & $43.2$ & $45.1$ & $52.7$ & $30.4$ & $\mathbf{66.5}$ & $55.4$ & $92.6$ & $55.4$ & $\underline{49.5}$ & $49.5$ & $39.7$ & $41.8$ \\
\rowcolor{mergehighlightblue} \hspace{5mm} \iftmerge{} & $\mathbf{60.7}$ & $\underline{46.0}$ & $88.2$ & $\underline{45.0}$ & $\underline{66.5}$ & $\mathbf{59.4}$ & $\mathbf{97.0}$ & $\underline{57.7}$ & $41.6$ & $\underline{50.8}$ & $79.3$ & $49.9$ \\
\bottomrule
\end{NiceTabular}
}
\vspace{1mm}
\end{table*}

\paragraph{Behavior changes during training}
An alternative to merging could be to abort training when a change in reasoning behavior is observed.
We implement an early stopping control (\emph{\iftabort{}}): every 5 training steps, we stop when the reasoning rate on $\mathcal{D}_{\mathrm{cal}}$ falls below $90\%$ and keep the last checkpoint before this threshold. Since \olmo{} does not change reasoning behavior, we exclude it from this comparison. Five of the remaining six settings cross this threshold within the first quarter of training (\cref{fig:training-trajectories}); the plotted checkpoints are too sparse to show the first few steps. Consequently, early stopping underperforms full \ift{} in 5 of 6 settings, with mean target-score deficits of $7.14\%$ on Rust and $11.32\%$ on summarization
  (\cref{tab:control-score-comparison}).

\paragraph{Causality of reasoning for performance}
Finally, we investigate whether the presence of reasoning, without any model training, recovers \oodbench{} performance. We force the \ift{} model to reason at inference time (\emph{+Force}) by preventing it from closing the reasoning trace without producing any content. As shown in \cref{tab:control-score-comparison}, this intervention exceeds the unmodified \ift{} target score in only 1 of 6 settings and remains below the \oodbench{} score of the untrained RLM in all six. Therefore, forcing a non-empty trace is insufficient to recover performance under our intervention.

\paragraph{Method variants}
For \openthinker{} on Rust, we compare our linear merging (Linear), Spherical Linear Interpolation (SLERP) \citep{slerp,mergekit}, and TIES \citep{ties}, shown in the leftmost panel of \cref{fig:openthinker-hyperparameters}. Target and \oodbench{} performance gains are $+3.79\%$ and $+0.60\%$ for Linear, $+3.93\%$ and $+2.28\%$ for SLERP, and $+3.23\%$ and $+1.52\%$ for TIES, respectively. While SLERP slightly outperforms linear merging, the results demonstrate that the overall trend is independent of merging technique.

In the second panel of \cref{fig:openthinker-hyperparameters}, we ablate the fine-tuning method, using full fine-tuning (Full FT) of all model weights.
Before merging, Full FT and LoRA reduce \oodbench{} performance by $-34.11\%$ and $-12.51\%$, respectively. Merging recovers these to $+0.86\%$ and $+0.60\%$, respectively. Again, the fine-tuning technique does not influence the trend.

\paragraph{Hyperparameter Ablation}
We ablate fine-tuning hyperparameters for the \ift{} stage in the three rightmost panels of \cref{fig:openthinker-hyperparameters}. Across seven configurations varying the number of epochs, learning rate, and batch size, our merging improves target task performance on average $2.87\%$. The remarkably stable performance may be caused by the adaptive reasoning ratio choice: Stronger nominal training settings generally select smaller merge ratios, $\alpha$ and $(\text{epochs}\times\eta\times16/B)$ have Spearman $\rho=-0.955$.


\begin{figure*}[t]
    \centering
    \includegraphics[width=1\linewidth]{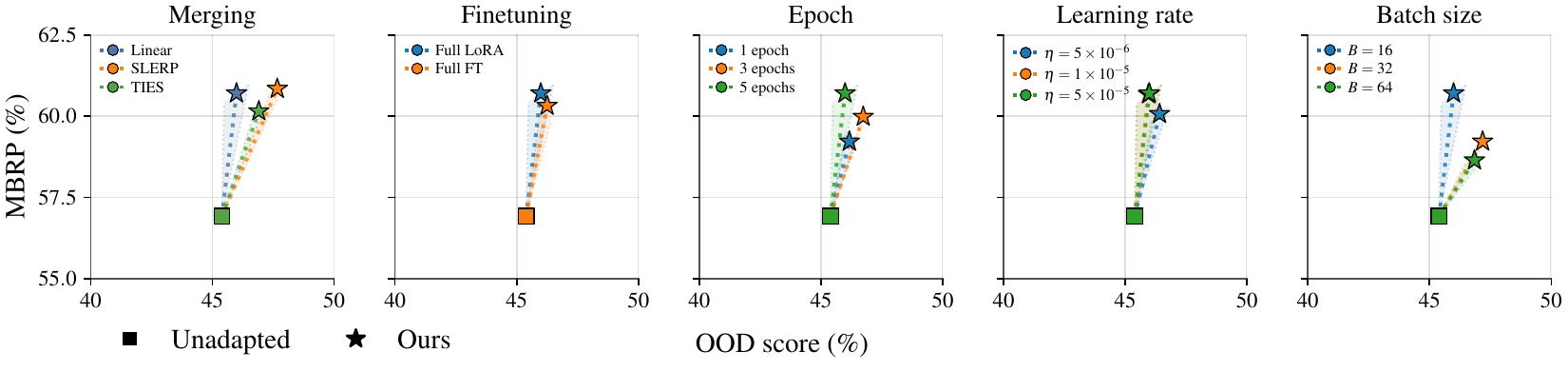}
    \caption{\openthinker{} ablations on Rust coding, from left to right: merging techniques, LoRA vs. full fine-tuning, training epochs, learning rate $\eta$, and batch size $B$. Across the seven hyperparameter configurations, the trend of preserved \oodbench{} performance and improved target performance holds.}
    \label{fig:openthinker-hyperparameters}
\vspace{-5mm}
\end{figure*}

\section{Related Work}
\label{sec:related-work}

We list prior work on adaptation of RLMs and model merging in general.

\paragraph{Training reasoning models with verifiers}
The standard technique for training and adapting reasoning models uses RLVR \citep{deepseekr1,olmo3,stojanovski2025reasoninggymreasoningenvironments}. This works well for tasks that permit such verifiers, but leaves out many domains, such as text summarization. Moreover, while this technique is well explored for bootstrapping reasoning model performance \citep{deepseekr1,qwq32b,olmo3}, continuous adaptation of RLMs to new tasks, as done in our setting, remains underexplored.

\paragraph{Training reasoning models without verifiers}
Verifier-free or verifier-light methods try to enable training reasoning models without reasoning traces by optimizing the likelihood of known answers directly \citep{verifree} or training teacher models to provide feedback and justifications for reference answers \citep{sdft}. These methods are close in motivation to our setting, but they are designed to elicit reasoning from models not previously trained for reasoning tasks.

\citet{lu2025onpolicydistillation} uses on-policy distillation from the base RLM on an unrelated instruction dataset to recover reasoning behavior after \ift{}. As we show in our experiments, our method is less expensive, does not introduce another hyperparameter for tuning, and does not depend on an external dataset.

\paragraph{Reasoning distillation and synthetic traces}
Assuming the availability of a stronger RLM on the desired task, reasoning distillation on synthetic traces has been proposed.
In reasoning distillation, for each task $\rvx$ in the dataset, a reasoning trace $\rvr$ and answer $\hat{\rvy}$ are obtained by sampling a stronger RLM \citep{deepseekr1,openthinker,qwen3}. Supervised fine-tuning is then performed on the new dataset with pairs $(\rvx, (\rvr, \hat{\rvy}))$.
However, this approach requires the availability of a stronger RLM on the desired task. In this work, we explore a setting where such a model cannot be obtained, since a stronger model is not available for any frontier training setups.

\paragraph{Model merging}
Model merging combines multiple checkpoints into a single model. Prior methods include simple weight averaging \citep{mergekit}, task arithmetic \citep{ilharco2022editing}, and sparse or sign-based variants such as TIES \citep{ties}. Merging is usually performed to combine task-specific skills or mitigate forgetting \citep{catastrophicforgetting,yang2024modelmerging,merge-transfer}. Our use is narrower: we merge an \ift{} checkpoint back with its own original reasoning checkpoint to trade off target-task adaptation against preservation of reasoning behavior. Recent work on tunable reasoning through model merging suggests that interpolation can control the degree of reasoning behavior in language models \citep{lan2025thinking}.

\vspace{-1mm}
\section{Discussion and Future Work}
\label{sec:limit}
\vspace{-1mm}

\paragraph{Inconsistent reasoning loss}
In our experiments, we observe that \openthinker{}, \qwen{} and \apriel{} lose their reasoning behavior in all settings, while \olmo{} never loses its reasoning behavior. We investigate this further and find that in one hyperparameter sweep setting, at learning rate $\eta=10^{-3}$ and 1 epoch, we observe reasoning loss for \olmo{} too, resulting in around $86\%$ reasoning. We find that for this setting, the norm of the weight update is $167.29$, more than 10 times the norms of our main experiments, $14.80$ and $14.82$. However, in our validation, this setting underperforms the hyperparameters chosen for our main experiments. This indicates interesting future work on exploring the relationship between task adaptation, weight updates, and reasoning loss more closely.


\paragraph{Lightweight adaptation on reasoning domains}
Our method provides a lightweight way to adapt reasoning models to new tasks by leveraging instruction-completion pairs, following the same paradigm as in the pre-RLM era.
However, more knowledge may provide great benefit on domains where the quality of reasoning itself is relevant to performance improvements, such as mathematics.
Indeed, in preliminary experiments, we applied our method to mathematical proofs, and did not observe significant target-task improvements. We believe this remains interesting future work.

\paragraph{Composability and continual learning}
Our results demonstrate how to adapt RLMs to a single task using IFT data while preventing changes in reasoning behavior or the loss of out-of-distribution knowledge. An interesting direction for future work is to explore the degree of composability enabled by this finding, as it potentially enables new approaches to continual learning.
\vspace{-1mm}

\section{Conclusion}
\label{sec:conclusion}
\vspace{-1mm}

We studied how to adapt RLMs with standard input-output pairs. We introduce a simple two-step pipeline to adapt RLMs using \ift{} data: first fine-tune the RLM with standard \ift{}, then merge the fine-tuned checkpoint with the untuned model. The merge ratio is selected using the reasoning rate on a target-task calibration set.
Across the eight evaluated model--task settings, this procedure improves the target score by an average of $5.56\%$ and $2.56\%$ on Rust coding and text summarization, respectively. Meanwhile, it preserves out-of-distribution performance, with average changes of only $+0.60\%$ and $-1.94\%$.
Our ablations show the same qualitative recovery across seven hyperparameter configurations, full fine-tuning, and three merging algorithms.
Our method operates without a verifier, expensive rollouts, or stronger teacher. More importantly, it is cheaper than the evaluated OPD and RL baselines by on average $22.3\%$ and $84.6\%$, respectively.

\section*{Acknowledgements}
This work has been done as part of the grant SAFEAI (Certified Safe, Fair and Robust Artificial Intelligence). The work has received funding from the Swiss State Secretariat for Education, Research and Innovation (SERI), contract no. MB22.00088.

\bibliography{references}
\bibliographystyle{iclr2027_conference}

\appendix
\onecolumn

\lstdefinestyle{conversation-turn}{
    language={},
    keywordstyle=\ttfamily,
    commentstyle=\ttfamily,
    basicstyle=\ttfamily\scriptsize,
    breaklines=true,
    breakatwhitespace=false,
    breakautoindent=false,
    breakindent=10pt,
    columns=fullflexible,
    numbers=none,
    frame=none,
    xleftmargin=0pt,
    framexleftmargin=0pt,
    numbersep=0pt,
    escapeinside={(*@}{@*)},
    mathescape=true,
    showstringspaces=false
}
\newcommand{\conversationturnlabel}[1]{%
    \noindent\hfill\textsf{\scriptsize #1}\par%
}
\newcommand{\conversationturnrule}{%
    \par\nointerlineskip\vspace{1mm}%
    \noindent\rule{\linewidth}{0.4pt}%
    \par\vspace{1mm}%
}

\section{Experimental Details and Ablations}
\label{app:experiments}

In this section, we provide additional details about the implementation, hyperparameters, and datasets.

\subsection{Dataset Curation}
\label{app:rust-details}

\paragraph{Rust coding}
In our study, the Rust corpus from \citet{oxen-rust-dataset,oxen-rust-grpo} is used for supervised fine-tuning and in-distribution evaluation. Each sample in the corpus consists of four fields: a task identifier, a natural-language Rust prompt that describes the target function, a Rust code implementation that solves the task, and a list of executable test cases. Consequently, the dataset provides complete function-level programming tasks paired with verification harnesses, enabling evaluation of generated code correctness.

Before using the dataset, we performed verification and filtering to improve data quality and ensure experimental reproducibility. For each code-test pair, we compiled the provided Rust implementation and executed its associated test suite ten times in our local environment. We retained only samples for which the implementation successfully compiled and passed the complete test suite in all ten runs. This repeated-execution protocol filters out examples whose correctness is unstable across executions, including cases affected by nondeterministic behavior or code-test pairs that only pass stochastically. We did not manually repair such cases; any sample that failed compilation or testing in at least one run was excluded from the final dataset.

Starting from the original 7{,}554 Rust tasks, this filtering process retained 7{,}511 valid samples. We then split the filtered dataset into 6{,}761 examples for supervised fine-tuning and 750 examples for validation. Each entry in the final split retains the same four-field structure: task identifier (\texttt{task\_id}), Rust prompt (\texttt{rust\_prompt}), Rust implementation (\texttt{rust\_code}), and the complete list of test cases (\texttt{rust\_test\_list}).

We confirm that the obtained datasets are disjoint from the MBRP dataset by checking for closest matches using cosine similarity.

\paragraph{Text summarization}
We sample 10{,}000 and 400 instances from the TL;DR set of \citet{stiennon2022learningsummarizehumanfeedback} for our training and calibration set respectively.
The core of the 10{,}000 samples is formed by 9{,}113 samples with at least 20 summary words and 150 post words, and 887 additional random samples to fill to 10{,}000.

\subsection{Details of Task Scoring}
\label{app:scores-details}

\paragraph{Rust coding}
For Rust coding, we use the test suite provided by the MBPP translation as scorer. The model output receives a score of 1 when all test cases pass, and 0 otherwise.

\paragraph{Text summarization}
For text summarization, we report \summeval, the average of the four metrics fluency, consistency, relevance, and coherence on the CNN/Daily Mail split of the \summeval dataset \citep{summeval}. Each metric is assigned a score between 1 and 5 by \gemini{} \citep{deepmind2025geminipro} based on clearly outlined criteria, shown in \cref{app:prompt}. We multiply these scores by 20 and report them as percentages, so a rating of 5 corresponds to $100\%$.

To confirm the validity of this LLM-as-a-judge setup, we compare the \gemini{} ratings with human annotations provided by \citet{summeval} and obtain a mean Spearman rank correlation coefficient of $\bar{r}_s=0.65$, $p<0.001$ across the four metrics, indicating a consistently high level of agreement.

\begin{table}[t]
\centering
\caption{\summeval correlations for \gemini.}
\label{tab:summeval-gemini3}
\begin{tabular}{lccc}
\toprule
Metric & Spearman $\rho$ & p-value \\
\midrule
Fluency & 0.6280 & $1.69\times10^{-182}$ \\
Relevance & 0.5917 & $1.66\times10^{-159}$ \\
Coherence & 0.7058 & $3.30\times10^{-253}$ \\
Consistency & 0.6793 & $2.89\times10^{-227}$ \\
\bottomrule
\end{tabular}
\end{table}

\subsection{Hyperparameters}
\label{app:ift-hyperparameters}

We run each training twice with seed 42 and 43.

When using LoRA, we train the query, key, value, and output projections in self-attention, as well as the gate, up, and down projections in the feed-forward network.
We tune the \ift{} hyperparameters separately for each model and dataset combination using a validation set as training target. \Cref{tab:ift-hyperparameters} reports the resulting configurations together with the baseline and ablation hyperparameters. The \iftopd{} rows only report the training parameters of the subsequent \opd{} training; the preceding \ift{} uses the corresponding \ift{} row.

During merge search, we sample 20 tokens plus the length of the longest end-of-reasoning marker to determine whether a model reasons.
To further speed up the merging search, calibration evaluation is performed using vLLM in eager loading mode (skipping model graphs) with a batch size of 128. All merging runs are performed on a single GPU.

During inference, the maximum token limit is 16,384 tokens for all benchmarks. We consistently use temperature 0.6 and top-p 0.95. For \openthinker, we use the default repetition penalty of 1.05.

\begin{table}[t]
\centering
\caption{Hyperparameters used for the main experiments, baselines, and ablations. Batch size denotes the effective training batch size; $r$ denotes LoRA rank. For RL (RLVR for Rust, RLAIF for text summarization), training length lists the validation-selected checkpoint steps for runs 1 / 2, from a budget of 100 updates per run.}
\label{tab:ift-hyperparameters}
\label{tab:baseline-ablation-hyperparameters}
\setlength{\tabcolsep}{3pt}
\resizebox{\linewidth}{!}{%
\begin{tabular}{lllcccc}
\toprule
Dataset & Model & Method & Training length & Learning rate & Batch size & Extra \\
\midrule
Rust & \openthinker{} & \ift{} & 5 epochs & $5 \times 10^{-5}$ & 16 & $r=32$ \\
Rust & \olmo{} & \ift{} & 3 epochs & $5 \times 10^{-5}$ & 16 & $r=32$ \\
Rust & \apriel{} & \ift{} & 5 epochs & $1 \times 10^{-4}$ & 64 & $r=32$ \\
Rust & \qwen{} & \ift{} & 5 epochs & $2 \times 10^{-4}$ & 16 & $r=64$ \\
Text sum. & \openthinker{} & \ift{} & 1 epoch & $5 \times 10^{-5}$ & 16 & $r=32$ \\
Text sum. & \olmo{} & \ift{} & 1 epoch & $1 \times 10^{-5}$ & 16 & $r=32$ \\
Text sum. & \apriel{} & \ift{} & 1 epoch & $1 \times 10^{-4}$ & 64 & $r=32$ \\
Text sum. & \qwen{} & \ift{} & 3 epochs & $2 \times 10^{-4}$ & 32 & $r=64$ \\
\midrule
All & All & \iftforce{} & \multicolumn{3}{c}{\ift{} config} & force first 10 tokens \\
\midrule
Rust & \openthinker{} & RL & 75 / 25 steps & $2 \times 10^{-5}$ & 64 & $r=32$ \\
Rust & \olmo{} & RL & 100 / 25 steps & $2 \times 10^{-5}$ & 64 & $r=32$ \\
Rust & \apriel{} & RL & 50 / 75 steps & $2 \times 10^{-5}$ & 64 & $r=32$ \\
Rust & \qwen{} & RL & 100 / 100 steps & $2 \times 10^{-5}$ & 64 & $r=32$ \\
Text sum. & \openthinker{} & RL & 100 / 100 steps & $1 \times 10^{-5}$ & 64 & $r=32$ \\
Text sum. & \olmo{} & RL & 100 / 100 steps & $1 \times 10^{-5}$ & 64 & $r=32$ \\
Text sum. & \apriel{} & RL & 100 / 100 steps & $1 \times 10^{-5}$ & 64 & $r=32$ \\
Text sum. & \qwen{} & RL & 100 / 100 steps & $1 \times 10^{-5}$ & 64 & $r=32$ \\
\midrule
Rust & \openthinker{} & \iftopd{} & 25 steps & $2 \times 10^{-4}$ & 8 & $r=64$ \\
Rust & \olmo{} & \iftopd{} & 25 steps & $2 \times 10^{-4}$ & 8 & $r=32$ \\
Rust & \apriel{} & \iftopd{} & 50 steps & $4 \times 10^{-4}$ & 4 & $r=32$ \\
Rust & \qwen{} & \iftopd{} & 25 steps & $2 \times 10^{-4}$ & 8 & $r=64$ \\
Text sum. & \openthinker{} & \iftopd{} & 25 steps & $2 \times 10^{-4}$ & 8 & $r=64$ \\
Text sum. & \olmo{} & \iftopd{} & 5 steps & $1 \times 10^{-5}$ & 8 & $r=32$ \\
Text sum. & \apriel{} & \iftopd{} & 50 steps & $4 \times 10^{-4}$ & 4 & $r=32$ \\
Text sum. & \qwen{} & \iftopd{} & 25 steps & $2 \times 10^{-4}$ & 8 & $r=64$ \\
\midrule
Rust & \openthinker{} & Full FT & 5 epochs & $5 \times 10^{-5}$ & 16 & all weights \\
\bottomrule
\end{tabular}
}
\end{table}

\subsection{RL Setup}
\label{app:rl-setup}
\label{app:rlaif-setup}

\paragraph{RLVR}
We use the same 6,761 Rust training problems used for \ift{}, each containing a problem statement, reference solution, and tests. RL discards the reference solution and validates outputs using the reference tests. We filter out prompts longer than 2,048 tokens.
Validation requires complete Rust code. We compile the result with \texttt{rustc} and then execute the reference tests. The reward is 1 only if every test passes and 0 otherwise. Compilation errors, timeouts, crashes, and incomplete answers receive a reward of 0.

We train each RLM using GRPO and LoRA adapters with rank 32, $\alpha=64$, and dropout 0. For each model, we run 100 updates with two seeds and a learning rate of $2\times10^{-5}$. Each update samples 64 completions: eight solutions for each of eight problems. For rollouts, the maximum model output is 8,192 tokens, the temperature is 0.8, and top-$p$ is 1.

During training, we evaluate eight completions for each of 64 fixed problems from the separate 750-problem calibration set at steps 0, 25, 50, 75, and 100. We select the checkpoint with the highest mean test-pass rate, breaking ties in favor of the earliest checkpoint.

\paragraph{RLAIF} For text summarization, we implement Reinforcement Learning from AI Feedback (RLAIF) using the frozen untrained RLM as an automated judge. We use the 10{,}000 training examples from the filtered Reddit TLDR split \citep{stiennon2022learningsummarizehumanfeedback}.

Each RLM is trained using GRPO \citep{grpo} with the DAPO loss objective and group reward scaling. As in \ift{}, we train LoRA adapters with rank $r=32$, $\alpha=64$, and dropout 0, applied to all attention projections (\texttt{q\_proj}, \texttt{k\_proj}, \texttt{v\_proj}, \texttt{o\_proj}) and feed-forward layers (\texttt{gate\_proj}, \texttt{up\_proj}, \texttt{down\_proj}). Training runs for 100 update steps across two random seeds (42 and 43) with a constant learning rate of $1 \times 10^{-5}$ (0 warmup steps) using the AdamW optimizer ($\beta_1=0.9$, $\beta_2=0.999$, $\epsilon=10^{-8}$, weight decay 0). The KL divergence coefficient is set to $\beta=0.05$. Each update step uses an effective batch size of 64 completions, sampling 8 completions for each of 8 prompts. Rollouts generate up to 8{,}192 tokens with temperature 0.8 and top-$p$ 1.0. Training uses bfloat16, masks truncated completions from the loss, and applies token-level truncated importance-sampling corrections to vLLM rollouts.

During rollouts, the policy receives the standard summarization prompt (\cref{prompt:text-summarization}) and is instructed to place its complete summary after the \texttt{<SUMMARY>} tag. Candidate summaries are extracted from the completions; if extraction fails or the output summary is empty, a reward of 0 is assigned without querying the judge. Across all four models and both random seeds, the step 100 checkpoint is used for evaluation.

To avoid reliance on stronger LLM judges, which are unavailable in our setting, each RLM acts as its own judge using its frozen base weights without adapter weights. The judge evaluates candidate summaries across the four independent \summeval{} dimensions: relevance, factual consistency, fluency, and coherence. Each dimension is queried in an independent context using the corresponding rubric prompt (\cref{prompt:relevance-evaluation,prompt:fluency-evaluation,prompt:consistency-evaluation,prompt:coherence-evaluation}). 

Judge generation uses temperature 0.6, top-$p$ 0.95, unrestricted top-$k$, repetition penalty 1.0, and a maximum completion length of 16{,}384 tokens to permit extensive reasoning traces by the judge before outputting its final score. If a judge generation is truncated or fails to parse a valid numeric score, the same prompt is tried again with the next seed, for at most ten total attempts, starting at seed 0. Scores are assigned as for the main evaluation. The unnormalized quality score $\bar{S}$ is the arithmetic mean of the four dimensions:
\begin{equation}
\bar{S} = \frac{1}{4} \left(S_{\text{relevance}} + S_{\text{consistency}} + S_{\text{fluency}} + S_{\text{coherence}}\right) \in [1, 5].
\end{equation}
The final scalar reward normalized to $[0, 1]$ passed to GRPO is:
\begin{equation}
R = \frac{\bar{S} - 1.0}{4.0}.
\end{equation}

\subsection{Detailed Scores}
\label{app:summeval-detailed}
\label{app:ood-detailed}

\paragraph{SummEval}
We provide the individual \summeval{} dimensions in \cref{tab:ood-benchmark-results}.
They highlight a trade-off between consistency and relevance that is not visible in the average. For the \openthinker{}, \qwen{}, and \apriel{}, all models whose reasoning behavior changes after \ift{}, direct \ift{} increases consistency but reduces relevance. Merging recovers relevance to near or above the base score, except for \qwen{} ($94.7$, $77.0$, and $99.5$, respectively), while keeping consistency above the corresponding base score ($63.1$, $56.1$, and $89.4$). Thus, the aggregate summarization gains after merging primarily combine recovered relevance with retained consistency gains.
For RLAIF, all four \summeval{} dimensions decline substantially for \openthinker{}, \qwen{}, and \apriel{}. \olmo{} is the exception, with its summarization dimensions remaining close to the unadapted model.

\paragraph{\oodbench}
We provide the individual \oodbench{} results in \cref{tab:ood-benchmark-results}.
The per-benchmark results show that forgetting is concentrated in particular model--benchmark combinations. For \openthinker{}, direct \ift{} reduces \mathbench{} from $82.8$ to $46.4$ after Rust training and from $84.8$ to $14.7$ after summarization training; merging restores the scores to $83.7$ and $83.3$, respectively. Similarly, for \apriel{} trained on summarization, \ift{} reduces \mathbench{} from $96.0$ to $83.3$ and \gpqabench{} from $57.6$ to $28.3$, while merging restores them to $95.6$ and $56.1$. By contrast, \olmo{} changes less across the individual OOD benchmarks, and merging leaves its \ift{} checkpoint unchanged because its reasoning behavior is retained.
RLAIF's individual OOD scores remain comparatively close to those of the unadapted models. In particular, \olmo{} remains close to the unadapted model across all three OOD benchmarks.

\begin{table*}[p]
\centering
\caption{Detailed out-of-domain and SummEval results. We observe that \gpqabench{} and \ifbench{} scores are stable for most models, in contrast to reasoning-heavy \mathbench{}. In text summarization, the strongest improved metric by \ift{} is consistency, while relevance decreases.}
\label{tab:ood-benchmark-results}
\small
\setlength{\tabcolsep}{3pt}
\resizebox{\linewidth}{!}{%
\begin{NiceTabular}{llrrrcrrrcrrrr}
\toprule
& & \multicolumn{3}{c}{Rust coding} & & \multicolumn{3}{c}{Text summarization} & & \multicolumn{4}{c}{SummEval} \\
\cmidrule(lr){3-5}\cmidrule(lr){7-9}\cmidrule(lr){11-14}
Model & Method & Math500 & IFBench & GPQA D. & & Math500 & IFBench & GPQA D. & & Fluency & Relevance & Coherence & Consistency \\
\midrule
\multirow{7}{*}{\rotatebox{90}{\openthinker{}}}
 & Unadapted & $82.8$ & $\underline{17.0}$ & $\underline{36.4}$ &  & $\underline{84.8}$ & $\underline{18.7}$ & $\underline{37.4}$ &  & $\underline{99.3}$ & $89.2$ & $\mathbf{97.2}$ & $49.4$ \\
 & RL & $\mathbf{86.4}$ & $\underline{19.0}$ & $\underline{36.6}$ &  & $\underline{84.9}$ & $17.5$ & $\mathbf{40.4}$ &  & $33.7$ & $31.8$ & $33.2$ & $24.8$ \\
 & IFT & $46.4$ & $\mathbf{21.7}$ & $\underline{30.6}$ &  & $14.7$ & $\underline{16.2}$ & $29.0$ &  & $\mathbf{99.5}$ & $83.3$ & $\underline{96.8}$ & $\mathbf{86.9}$ \\
 & \quad +OPD & $81.3$ & $\underline{18.3}$ & $\underline{35.9}$ &  & $83.3$ & $\mathbf{21.3}$ & $\underline{38.9}$ &  & $80.8$ & $79.2$ & $77.1$ & $61.9$ \\
 & \quad \rowcolor{mergehighlightblue} \iftmerge{} & $83.7$ & $\underline{19.7}$ & $\underline{34.6}$ &  & $83.3$ & $\underline{16.8}$ & $\underline{34.8}$ &  & $\underline{99.5}$ & $\mathbf{94.7}$ & $95.3$ & $63.1$ \\
 & \quad +Abort & $84.7$ & $\underline{20.3}$ & $\mathbf{38.4}$ &  & $\mathbf{86.9}$ & $\underline{18.8}$ & $\underline{35.1}$ &  & $97.8$ & $89.7$ & $94.3$ & $51.9$ \\
 & \quad +Force & $82.4$ & $15.7$ & $\underline{37.4}$ &  & $44.6$ & $11.5$ & $\underline{35.1}$ &  & $59.6$ & $46.8$ & $55.8$ & $48.7$ \\
\midrule
\multirow{7}{*}{\rotatebox{90}{\qwen{}}}
 & Unadapted & $\mathbf{92.8}$ & $\underline{12.7}$ & $\underline{47.5}$ &  & $\mathbf{93.8}$ & $\underline{14.0}$ & $\underline{48.0}$ &  & $\underline{99.3}$ & $84.8$ & $83.7$ & $45.4$ \\
 & RL & $\underline{92.6}$ & $\underline{13.7}$ & $\underline{47.7}$ &  & $\underline{92.3}$ & $\underline{13.0}$ & $\mathbf{51.5}$ &  & $48.7$ & $42.5$ & $43.1$ & $28.8$ \\
 & IFT & $89.3$ & $\underline{14.7}$ & $41.9$ &  & $88.2$ & $8.7$ & $41.2$ &  & $97.0$ & $67.7$ & $89.5$ & $\mathbf{70.8}$ \\
 & \quad +OPD & $\underline{90.8}$ & $12.2$ & $\underline{47.0}$ &  & $\underline{92.5}$ & $\underline{13.8}$ & $\underline{49.0}$ &  & $\mathbf{99.5}$ & $\mathbf{88.7}$ & $\mathbf{91.0}$ & $55.2$ \\
 & \quad \rowcolor{mergehighlightblue} \iftmerge{} & $\underline{92.6}$ & $\mathbf{16.2}$ & $43.7$ &  & $91.6$ & $\underline{12.8}$ & $\underline{45.2}$ &  & $98.0$ & $77.0$ & $86.0$ & $56.1$ \\
 & \quad +Abort & $\underline{92.6}$ & $\underline{13.7}$ & $\mathbf{50.3}$ &  & $\underline{92.5}$ & $\mathbf{15.0}$ & $\underline{46.5}$ &  & $92.2$ & $69.7$ & $81.9$ & $65.4$ \\
 & \quad +Force & $90.1$ & $\underline{13.8}$ & $\underline{44.7}$ &  & $90.4$ & $9.2$ & $25.8$ &  & $56.0$ & $29.5$ & $42.3$ & $31.1$ \\
\midrule
\multirow{7}{*}{\rotatebox{90}{\olmo{}}}
 & Unadapted & $\underline{94.0}$ & $\underline{40.3}$ & $\underline{45.5}$ &  & $\underline{94.6}$ & $\mathbf{43.3}$ & $\underline{47.0}$ &  & $\mathbf{99.9}$ & $95.5$ & $95.0$ & $65.4$ \\
 & RL & $\underline{94.2}$ & $\mathbf{44.2}$ & $\mathbf{51.0}$ &  & $\underline{94.2}$ & $\underline{40.2}$ & $\mathbf{49.7}$ &  & $99.7$ & $95.4$ & $\underline{95.7}$ & $64.4$ \\
 & IFT & $\mathbf{95.2}$ & $35.8$ & $\underline{47.2}$ &  & $\underline{94.2}$ & $36.5$ & $\underline{45.2}$ &  & $99.5$ & $\underline{96.1}$ & $\underline{96.4}$ & $\mathbf{77.7}$ \\
 & \quad +OPD & $93.5$ & $38.7$ & $\underline{48.7}$ &  & $\mathbf{95.4}$ & $37.5$ & $\underline{48.2}$ &  & $\underline{99.8}$ & $\mathbf{96.6}$ & $\mathbf{96.8}$ & $70.4$ \\
 & \quad \rowcolor{mergehighlightblue} \iftmerge{} & $\mathbf{95.2}$ & $35.8$ & $\underline{47.2}$ &  & $\underline{94.2}$ & $36.5$ & $\underline{45.2}$ &  & $99.5$ & $\underline{96.1}$ & $\underline{96.4}$ & $\mathbf{77.7}$ \\
 & \quad +Abort & $\mathbf{95.2}$ & $35.8$ & $\underline{47.2}$ &  & $\underline{94.2}$ & $36.5$ & $\underline{45.2}$ &  & $99.5$ & $\underline{96.1}$ & $\underline{96.4}$ & $\mathbf{77.7}$ \\
 & \quad +Force & $\mathbf{95.2}$ & $35.8$ & $\underline{47.2}$ &  & $\underline{94.2}$ & $36.5$ & $\underline{45.2}$ &  & $99.5$ & $\underline{96.1}$ & $\underline{96.4}$ & $\mathbf{77.7}$ \\
\midrule
\multirow{7}{*}{\rotatebox{90}{\apriel{}}}
 & Unadapted & $\underline{95.6}$ & $\underline{22.0}$ & $\underline{53.0}$ &  & $\mathbf{96.0}$ & $\underline{21.7}$ & $\mathbf{57.6}$ &  & $\mathbf{100.0}$ & $99.2$ & $\mathbf{99.4}$ & $84.0$ \\
 & RL & $\mathbf{96.6}$ & $\underline{24.0}$ & $\underline{54.3}$ &  & $\underline{95.6}$ & $\mathbf{24.2}$ & $\underline{57.1}$ &  & $85.3$ & $85.0$ & $84.5$ & $72.6$ \\
 & IFT & $94.6$ & $\underline{22.8}$ & $52.0$ &  & $83.3$ & $18.7$ & $28.3$ &  & $99.6$ & $83.6$ & $97.7$ & $\mathbf{95.2}$ \\
 & \quad +OPD & $94.7$ & $21.0$ & $\underline{54.8}$ &  & $\underline{94.6}$ & $16.7$ & $\underline{50.5}$ &  & $98.7$ & $\underline{99.5}$ & $98.9$ & $91.9$ \\
 & \quad \rowcolor{mergehighlightblue} \iftmerge{} & $95.3$ & $\underline{24.2}$ & $\mathbf{58.6}$ &  & $\underline{95.6}$ & $\underline{21.3}$ & $\underline{56.1}$ &  & $\underline{100.0}$ & $\mathbf{99.5}$ & $\underline{99.3}$ & $89.4$ \\
 & \quad +Abort & $\underline{95.6}$ & $\mathbf{24.5}$ & $\underline{56.6}$ &  & $\underline{95.6}$ & $\underline{21.3}$ & $\underline{52.8}$ &  & $72.8$ & $72.8$ & $71.9$ & $71.2$ \\
 & \quad +Force & $93.3$ & $20.7$ & $52.3$ &  & $\underline{94.9}$ & $\underline{20.7}$ & $\underline{50.8}$ &  & $95.0$ & $93.8$ & $91.1$ & $90.6$ \\
\bottomrule
\end{NiceTabular}
}
\end{table*}

\subsection{Reasoning Length and Performance}
\label{app:reasoning-length}

We additionally investigate whether producing a longer reasoning trace is associated with better performance. We count tokens before the model's native end-of-reasoning marker using its native tokenizer, excluding the reasoning markers themselves. Empty traces have length zero. We first average over examples and evaluation seeds, and then equally weight available training runs at each merge ratio or training checkpoint. \Cref{fig:reasoning-lengths} shows the resulting trajectories.

\begin{figure}[p]
\centering
\includegraphics[width=.82\linewidth]{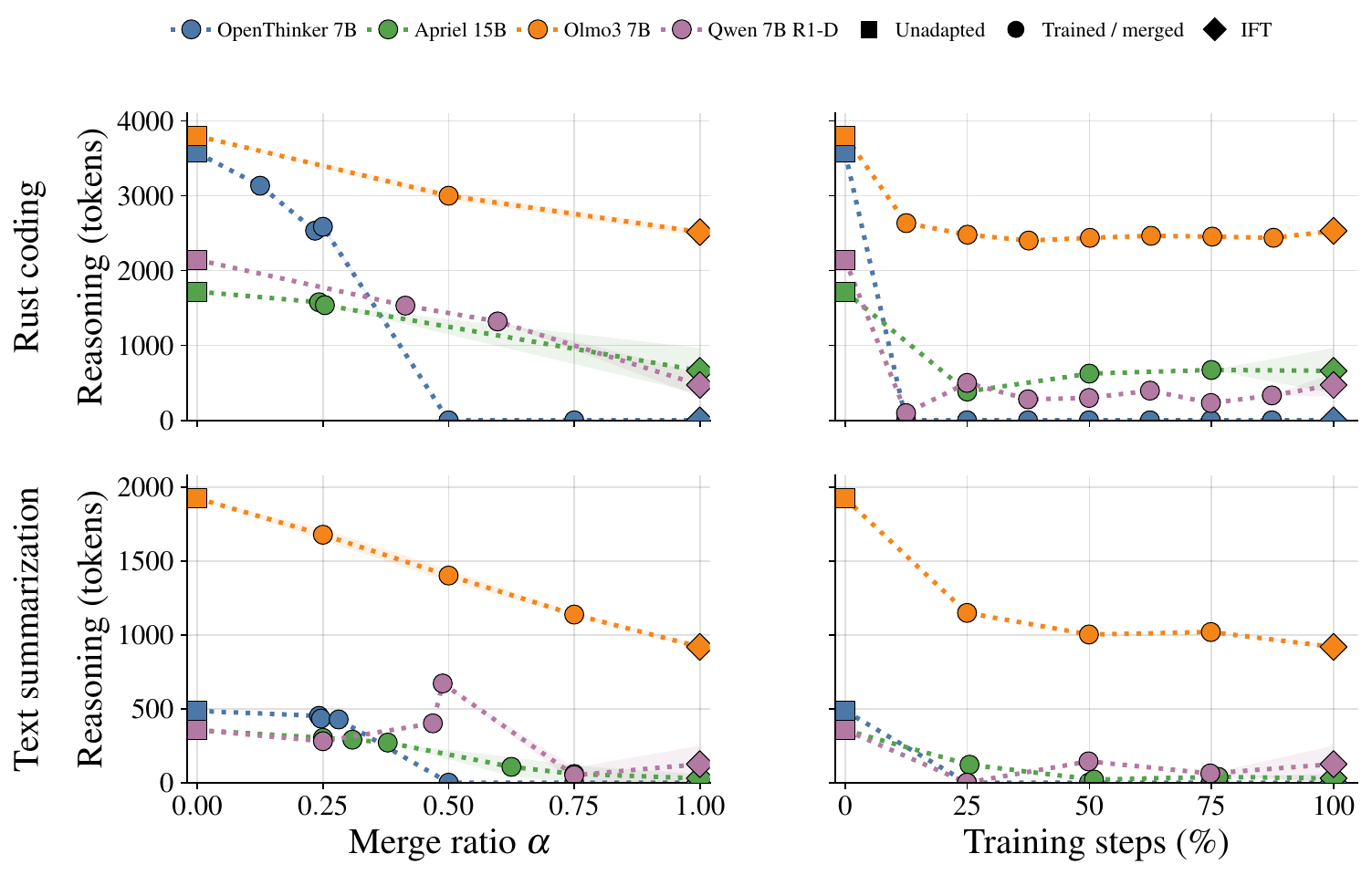}
\caption{Mean target-task reasoning length across merge ratios (left) and \ift{} training progress (right) indicate that all models experience a slight decrease in reasoning length. We can not determine a relationship to task performance beyond reasoning presence or absence.}
\label{fig:reasoning-lengths}
\end{figure}

Mean trace length confounds two distinct quantities: how often the model reasons and how long it reasons when it does. This distinction matters because the loss of reasoning behavior is already associated with capability loss in our experiments. To focus on length rather than presence, we restrict the analysis to points with at least $90\%$ target-task reasoning and divide mean length by the non-empty-trace rate. This gives the mean length conditional on a non-empty trace. The retained points still span substantial ranges: 1{,}321--3{,}799 tokens for the Rust merge trajectories, 311--3{,}799 for Rust training, 276--1{,}923 for text merging, and 355--1{,}923 for text training.

\Cref{tab:reasoning-length-correlation} reports exploratory Spearman correlations pooled across models. We find no evidence that longer traces improve target-task performance: the four correlations are small ($\rho=-0.17$--$0.07$) and none is significant (Holm-adjusted $p=1.00$ for all four). For example, \olmo{} on text summarization, mean non-empty reasoning length falls from 1{,}923 to 1{,}137 tokens between $\alpha=0$ and $\alpha=0.75$, while \summeval increases from $89.0$ to $92.4$ and reasoning remains at $100\%$. For \qwen{} on Rust, length falls from 2{,}144 to approximately 1{,}321 tokens by $\alpha=0.60$, while MBRP rises from $30.6$ to $44.2$ and reasoning remains above $99\%$.

Some correlations with \oodbench{} are positive, especially for the text trajectories. These do not establish a benefit from longer reasoning: merge ratio and training progress jointly change trace length, task adaptation, and forgetting, so length is not independently manipulated. Moreover, neither text correlation remains significant after Holm correction across the eight exploratory tests in \cref{tab:reasoning-length-correlation}. We therefore interpret the results as evidence that preserving the \emph{presence} of reasoning is informative, whereas the available experiments provide no evidence that making an already non-empty trace longer improves target-task performance.

\begin{table}[t]
\centering
\caption{Spearman correlations ($\rho$) between mean reasoning length conditional on a non-empty trace and performance on points with at least $90\%$ target-task reasoning. Holm-adjusted $p$-values across all eight tests are not significant.}
\label{tab:reasoning-length-correlation}
\small
\begin{tabular}{llrrrrr}
\toprule
&  && \multicolumn{2}{c}{Target score} & \multicolumn{2}{c}{\oodbench{}} \\
\cmidrule(lr){4-5}\cmidrule(lr){6-7}
Task & Trajectory & $n$ & $\rho$ & $p$ & $\rho$ & $p$ \\
\midrule
\multirow{2}{*}{Rust} & Merge ratios & 13 & -0.011 & 1.000 & -0.110 & 1.000 \\
& Training checkpoints & 15 & 0.071 & 1.000 & 0.521 & 0.324 \\
\multirow{2}{*}{Text sum.}& Merge ratios & 17 & -0.167 & 1.000 & 0.343 & 1.000 \\
 & Training checkpoints & 8 & -0.024 & 1.000 & 0.881 & 0.031 \\
\bottomrule
\end{tabular}
\end{table}

\subsection{Further Baseline Experiments}
\label{appsec:more-baseline}

We attempted to reproduce two popular related methods that are also applicable to our setting of training RLMs using only standard \ift{} data.

The first method, VeriFree \citep{verifree}, requires substantial resources. Based on our initial experiments, we estimated that a single training run would require 4 H200 GPUs for \SI{70}{\hour}. Given this cost, which applies to each step of a preliminary hyperparameter search for fair comparison, we consider a comparison infeasible.

Second, we made extensive attempts to reproduce \textsc{SDFT} \citep{sdft} on \olmo{} for Rust coding.
We performed 26 runs with the following \ift{} settings: epochs $\in \{1,2,3,5\}$, learning rate $\eta \in \{1\times 10^{-4}, 5\times 10^{-5}, 1\times 10^{-5}, 5\times 10^{-6}\}$, batch size $B \in \{16,32,64\}$, and EMA coefficient $\alpha \in \{0.01,0.02,0.05\}$. In order to fit into the GPU memory, matching our RL setup, we set the maximum token length for rollouts to $4,096$ and discarded training samples exceeding this budget, leaving roughly one-third of the examples. The best-performing run used two epochs, a learning rate of $5\times 10^{-6}$, a batch size of $64$, and EMA $\alpha=0.05$, and completed in \SI{4.5}{\hour} using four H200 GPUs.
The best performance we could achieve with \textsc{SDFT} was $41.4\%$, a decrease from the untuned performance of $46.2\%$, while performance on \oodbench{} also slightly decreased from $95.5\%$ to $93.1\%$.
As such, despite our extensive reproduction efforts, we did not include the above related methods in the direct comparison in our experiments.

\subsection{Runtime Details}
\label{app:runtime-details}

We report the runtime breakdown for the total runtime (Total) of our method in \cref{tab:runtime-breakdown}, generated from the same per-run runtimes as \cref{tab:main-score-cost}. The total runtime consists of the \ift{} training (\ift), model merge operations (Merge), token sampling on the calibration dataset (Cal.), model merge loading to prepare sampling (Load), and auxiliary overhead (Other).

\begin{table}[t]
\centering
\caption{Runtime breakdown for our method in minutes. Cal. reports separately measured calibration evaluation; Load and Other report model loading and residual overhead, respectively.}
\label{tab:runtime-breakdown}
\setlength{\tabcolsep}{4pt}
\begin{tabular}{llrrrrrr}
\toprule
Task & Model & \ift{} & Merge & Cal. & Load & Other & Total \\
\midrule
Rust & \openthinker{} 7B & 46.1 & 1.32 & 1.36 & 1.51 & 0.43 & 51.6 \\
Rust & \qwen{} 7B & 69.8 & 0.03 & 2.38 & 1.33 & 0.30 & 74.1 \\
Rust & \olmo{} 7B & 31.6 & 0.35 & 0.24 & 1.20 & 0.35 & 34.1 \\
Rust & \apriel{} 15B & 159.0 & 0.06 & 4.83 & 1.52 & 0.41 & 166.2 \\
Text sum. & \openthinker{} 7B & 14.7 & 0.87 & 1.05 & 1.81 & 0.38 & 19.6 \\
Text sum. & \qwen{} 7B & 66.5 & 0.02 & 1.71 & 1.28 & 0.22 & 69.9 \\
Text sum. & \olmo{} 7B & 16.9 & 0.38 & 0.19 & 1.28 & 0.33 & 19.4 \\
Text sum. & \apriel{} 15B & 33.5 & 1.70 & 2.12 & 1.36 & 0.65 & 40.8 \\
\midrule
Average & Rust & 76.6 & 0.44 & 2.20 & 1.39 & 0.37 & 81.5 \\
Average & Text sum. & 32.9 & 0.74 & 1.26 & 1.43 & 0.39 & 37.4 \\
\midrule
Average & All & 54.8 & 0.59 & 1.73 & 1.41 & 0.38 & 59.4 \\
\bottomrule
\end{tabular}
\end{table}

%

\section{Case Study}
\label{app:rust-case-study-outputs}

In this section, we provide detailed outputs for an MBRP example where the model is tasked with producing code that finds the correct insertion point in a sorted array. The unadapted model \apriel{}, shown in \cref{fig:rust-case-study-base}, reasons correctly and produces sensible code but fails to recognize a type mismatch. It initializes the running variable \texttt{low} as \texttt{0} and \texttt{high} from the length method \texttt{a.len()}. The length method returns the type \texttt{usize} (unsigned integer), so \texttt{high} is inferred to have type \texttt{usize}. Meanwhile, numeric literals in Rust do not have a fixed type and can be unsigned or signed depending on context. Since \texttt{low} interacts with \texttt{high}, it is also inferred to have type \texttt{usize} and thus mismatches the target return type \texttt{isize}.

The \ift{} model, shown in \cref{fig:rust-case-study-sft}, gives a correct, direct answer without reasoning by immediately casting the output of \texttt{a.len()} to \texttt{isize}, causing both \texttt{low} and \texttt{high} to be inferred to have the correct type.

Finally, the merged model, shown in \cref{fig:rust-case-study-merge}, reasons like the unadapted model but recognizes the potential type conflict early (as highlighted in the reasoning trace in \textcolor{blue}{blue}). As a result, it produces code similar to that of the unadapted model with a crucial type cast to \texttt{isize} for the return value (highlighted in \textcolor{red}{red}).

Note that \apriel{} does not generate explicit thinking markers, but only \texttt{[BEGIN FINAL RESPONSE]} to mark the final answer. We insert thinking tokens (\texttt{[BEGIN THOUGHT]} and \texttt{[END THOUGHT]}) for clarity.

\begin{figure}
\centering
\begin{tcolorbox}[colback=white,colframe=black,boxrule=0.4pt,arc=0pt,left=1mm,right=1mm,top=1mm,bottom=1mm]
\conversationturnlabel{User}
\begin{lstlisting}[style=conversation-turn]
Write a function to locate the left insertion point for a specified value in sorted order. The signature of the function is: 'fn left_insertion(a: Vec<isize>, x: isize) -> isize'
\end{lstlisting}

\conversationturnrule
\conversationturnlabel{Assistant}
\begin{lstlisting}[style=conversation-turn,basicstyle=\ttfamily\tiny]
[BEGIN THOUGHT]
Okay, I need to write a Rust function called left_insertion that takes a vector of isize and a value x, and returns the left insertion point. The left insertion point is the index where the element x should be inserted to maintain the sorted order, such that all elements before are less than x.

First, I remember that in Python, the bisect module has bisect_left which does exactly this. So the approach here is similar. The vector is sorted, so we can perform a binary search.

Let me think about the steps. The function needs to find the first position where a[i] >= x. If all elements are smaller, then the insertion point is the length of the vector.

So the binary search will have low and high pointers. Initialize low to 0 and high to a.len(). Then, while low is less than high, find the mid point. If a[mid] is less than x, then the insertion point must be to the right of mid, so set low to mid + 1. Otherwise, set high to mid. When the loop ends, low should be the insertion point.
Wait, let me test this logic with an example. Suppose the vector is [1,3,5,7], and x is 4. The mid points would be calculated, and the loop would adjust low and high until it finds the correct position. Let's see:
Initial low=0, high=4. Mid is 2. a[2] is 5, which is greater than 4. So high becomes 2. Now low is 0, high 2. Mid is 1. a[1] is 3, which is less than 4. So low becomes 2. Now low equals high, loop ends. Return 2, which is correct.
Another example: x is 0. The insertion point is 0. The loop would check mid 2 (5), which is greater, so high becomes 2. Then mid 1 (3), high becomes 1. Then mid 0 (1), high becomes 0. Loop ends, return 0.
Another case: x is 8. The insertion point is 4. The loop would proceed with low starting at 0, high 4. Mid 2 (5 <8), low becomes 3. Mid (3+4)/2=3.5, which is 3. a[3] is7 <8, so low becomes4. Now low equals high, return 4.

So the algorithm seems correct.

Now, implementing this in Rust. The function signature is given as fn left_insertion(a: Vec<isize>, x: isize) -> isize.

So the code would be:

Initialize low as 0, high as a.len(). Then loop while low < high. Compute mid as (low + high) / 2. But in Rust, integer division is done with /, and for even and odd numbers, it truncates towards zero. So (low + high) / 2 is correct.

Compare a[mid] with x. If a[mid] < x, then set low = mid +1. Else, set high = mid. At the end, return low.

But wait, what if the vector is empty? Then a.len() is 0, and the loop doesn't run, returns 0, which is correct.

So the code would look like:

[... copy of code below ...]

Yes, that should work. Let me check the examples again. For the vector [1,3,5,7], x=4, the code returns 2. For x=0, returns 0. For x=8, returns 4. All correct.

So this should be the correct implementation.
[END THOUGHT]
[BEGIN FINAL RESPONSE]
```rust
fn left_insertion(a: Vec<isize>, x: isize) -> isize {
    let mut low = 0;
    let mut high = a.len();
    while low < high {
        let mid = (low + high) / 2;
        if a[mid] < x {
            low = mid + 1;
        } else {
            high = mid;
        }
    }
    low
}
```
[END FINAL RESPONSE]
\end{lstlisting}

\conversationturnrule
\conversationturnlabel{Compiler}
\begin{lstlisting}[style=conversation-turn,basicstyle=\ttfamily\tiny]
error[E0308]: mismatched types
  --> /var/folders/tc/ybwwqpzd0774whvhgf_b077m0000gn/T/apriel_diag_292_t0fq4h2a/case.rs:12:5
   |
 1 | fn left_insertion(a: Vec<isize>, x: isize) -> isize {
   |                                               ----- expected `isize` because of return type
...
12 |     low
   |     ^^^ expected `isize`, found `usize`
\end{lstlisting}
\end{tcolorbox}
\captionof{figure}{The non-tuned \apriel{} model produces a non-empty reasoning trace but fails to produce compiling Rust code because it does not recognize that the implicit type of variable \texttt{low} is \texttt{usize}, which conflicts with the target return type \texttt{isize}.}
\label{fig:rust-case-study-base}
\end{figure}

\begin{figure}
\centering
\begin{tcolorbox}[colback=white,colframe=black,boxrule=0.4pt,arc=0pt,left=1mm,right=1mm,top=1mm,bottom=1mm]
\conversationturnlabel{User}
\begin{lstlisting}[style=conversation-turn]
Write a function to locate the left insertion point for a specified value in sorted order. The signature of the function is: 'fn left_insertion(a: Vec<isize>, x: isize) -> isize'
\end{lstlisting}

\conversationturnrule
\conversationturnlabel{Assistant}
\begin{lstlisting}[style=conversation-turn]
[BEGIN THOUGHT]
[END THOUGHT]
[BEGIN FINAL RESPONSE]
```rust
fn left_insertion(a: Vec<isize>, x: isize) -> isize {
    let mut low = 0;
    let mut high = (*@\color{red}{a.len() as isize}@*);

    while low < high {
        let mid = low + (high - low) / 2;
        if a[mid as usize] < x {
            low = mid + 1;
        } else {
            high = mid;
        }
    }

    low
}
```
[END FINAL RESPONSE]
\end{lstlisting}
\end{tcolorbox}
\captionof{figure}{The \ift{} model emits an empty reasoning trace and then a direct correct solution by immediately casting the running variables to the target type \texttt{isize} (highlighted in \textcolor{red}{red}).}
\label{fig:rust-case-study-sft}
\end{figure}
\begin{figure}
%
%
%
%
%
    
\centering
\begin{tcolorbox}[colback=white,colframe=black,boxrule=0.4pt,arc=0pt,left=1mm,right=1mm,top=1mm,bottom=1mm]
\conversationturnlabel{User}
\begin{lstlisting}[style=conversation-turn]
Write a function to locate the left insertion point for a specified value in sorted order. The signature of the function is: 'fn left_insertion(a: Vec<isize>, x: isize) -> isize'
\end{lstlisting}

\conversationturnrule
\conversationturnlabel{Assistant}
\begin{lstlisting}[style=conversation-turn,basicstyle=\ttfamily\tiny]
[BEGIN THOUGHT]
Okay, I need to write a Rust function called left_insertion that finds the left insertion point for a value x in a sorted vector a. The function should return the index where x should be inserted to maintain the sorted order. Let me think about how to approach this.

First, I remember that the left insertion point is the first position where the element is greater than or equal to x. So, for example, if the vector is [1,3,5,7] and x is 4, the left insertion point would be 2 because 5 is the first element greater than 4.  [...]

So the binary search approach seems to work. Now, how to implement this in Rust.

The function signature is given as fn left_insertion(a: Vec<isize>, x: isize) -> isize. So the parameters are a vector of isize and x is isize. The return is isize, which is the index.

Implementing the binary search:

Initialize low as 0, high as a.len() as isize. (*@\color{blue}{Wait, but in Rust, the vector's len() is usize. So when we do operations, we need to handle the types. Let's see. Let's use usize for low and high. Because the insertion point can't be larger than the vector's length, which is a usize.}@*)

So, low starts at 0, high starts at a.len(). Then, while low < high, compute mid. Mid can be (low + high) / 2. But in Rust, adding two usize can cause overflow, but since the vector's length is a valid usize, and low and high are within that, it's safe here.

Wait, but in the case where the vector is empty, a.len() is 0, so high is 0, and the loop doesn't run. Then return 0, which is correct.

So, code steps:

[... copy of code below ...]

Wait, but in Rust, the vector is passed as a Vec<isize>. So when accessing a[mid], if mid is within the vector's bounds, it's okay. But in the case where the vector is empty, the loop doesn't run, and low is 0, which is correct.

Testing the code with the examples:
Example 1: a = vec![1,3,5,7], x=4. The code should return 2. Let's see:
low=0, high=4. mid=2. a[2] is 5 >=4, so high=2. Now low=0, high=2. mid=1. a[1] is 3 <4, so low=2. Now low=2, high=2. Loop ends. Return 2. Correct.

Another example: x=0. The code returns 0. Correct.
Another example: x=8. The code returns 4. Correct.
What about when the element is present in the vector? For example, a = vec![2,4,4,4,6], x=4. The left insertion point is 1. Let's see:
low=0, high=5. mid=2. a[2] is 4 >=4, so high=2. Now low=0, high=2. mid=1. a[1] is 4 >=4, so high=1. Now low=0, high=1. mid=0. a[0] is 2 <4, so low=1. Now low=1, high=1. Return 1. Correct.

So the code seems to handle all cases correctly.

Therefore, the implementation should be as written.

[END THOUGHT]
[BEGIN FINAL RESPONSE]
```rust
fn left_insertion(a: Vec<isize>, x: isize) -> isize {
    let mut low = 0;
    let mut high = a.len();
    while low < high {
        let mid = (low + high) / 2;
        if a[mid] < x {
            low = mid + 1;
        } else {
            high = mid;
        }
    }
    (*@\color{red}{low as isize}@*)
}
```
[END FINAL RESPONSE]
\end{lstlisting}
\end{tcolorbox}
\caption{The model merged with our method produces a non-empty reasoning trace and correctly recognizes the implicit types of \texttt{low} and \texttt{high} and the potential conflict with the function signature (highlighted in \textcolor{blue}{blue}). It thus inserts a crucial cast at the return position (highlighted in \textcolor{red}{red}).}
\label{fig:rust-case-study-merge}
\end{figure}

\section{Prompts}
\label{app:prompt}

We use one task prompt for Rust code generation, one task prompt for text summarization, and four judge prompts for text summarization evaluation and RLAIF training.
\Cref{prompt:rust-code-generation} shows the Rust prompt used for both \ift{} training and MBRP evaluation.
\Cref{prompt:text-summarization} shows the text summarization prompt used for both \ift{} training and RLAIF rollouts on the TLDR split, as well as evaluation on the CNN/Daily Mail split.
For the LLM-based text summarization metrics, as well as the reward computation in RLAIF training, the models are evaluated separately for each criterion: \cref{prompt:relevance-evaluation} for relevance, \cref{prompt:fluency-evaluation} for fluency, \cref{prompt:consistency-evaluation} for factual consistency, and \cref{prompt:coherence-evaluation} for coherence.

\begin{figure}[p]
\centering
\begin{tcolorbox}[colback=white,colframe=black,boxrule=0.4pt,arc=0pt,left=1mm,right=1mm,top=1mm,bottom=1mm]
\conversationturnlabel{User}
\begin{lstlisting}[style=conversation-turn]
You are a helpful coding assistant producing high-quality Rust code.
Strictly follow the instruction below to complete the function specified in the
instruction.
Your response should include all dependencies, headers and function declaration
to be directly usable (even for the ones seen in the given part).
You should NOT call or test the function in your response.
Output your complete implementation in a single code block wrapped in triple
backticks with `rust` specified, like this:
```rust
// your function here
```
Instruction:
{{RUST_PROMPT}}
\end{lstlisting}
\end{tcolorbox}
\caption{Prompt used for both training and evaluation in the Rust code generation task. The task instruction varies per example and is inserted as \texttt{\{\{RUST\_PROMPT\}\}}.}
\label{prompt:rust-code-generation}
\end{figure}

\begin{figure}[p]
\centering
\begin{tcolorbox}[colback=white,colframe=black,boxrule=0.4pt,arc=0pt,left=1mm,right=1mm,top=1mm,bottom=1mm]
\conversationturnlabel{User}
\begin{lstlisting}[style=conversation-turn]
{{PROBLEM}}
Please reason step by step, and put your final answer within \boxed{}.
\end{lstlisting}
\end{tcolorbox}
\caption{Prompt used for MATH-500 evaluation. The problem is inserted as \texttt{\{\{PROBLEM\}\}}.}
\label{prompt:math500-evaluation}
\end{figure}

\begin{figure}[p]
\centering
\begin{tcolorbox}[colback=white,colframe=black,boxrule=0.4pt,arc=0pt,left=1mm,right=1mm,top=1mm,bottom=1mm]
\conversationturnlabel{User}
\begin{lstlisting}[style=conversation-turn]
You are an expert in writing summarization.
Your task is to read the following Article and write a summary about it.

Output your complete summary after the <SUMMARY> tag, like this:
<SUMMARY>
// your summary here

Article:
{{ARTICLE}}
\end{lstlisting}
\end{tcolorbox}
\caption{Prompt used for both training and evaluation in the text summarization task. The article is inserted as \texttt{\{\{ARTICLE\}\}}.}
\label{prompt:text-summarization}
\end{figure}

\begin{figure}[p]
\centering
\begin{tcolorbox}[colback=white,colframe=black,boxrule=0.4pt,arc=0pt,left=1mm,right=1mm,top=1mm,bottom=1mm]
\conversationturnlabel{User}
\begin{lstlisting}[style=conversation-turn]
You are a helpful assistant in evaluating the quality of a summary.
You will be given a news article and a summary written for that article.
Your task is to evaluate the *relevance* of the summary.

Definition of Relevance:
Relevance measures how well the summary captures the important information from
the article. A relevant summary includes the most important main idea of the
article and avoids redundancies and minor, trivial, or unrelated details.

Evaluation Criteria (Relevance: 1-100)

80-100 -- The summary captures the key idea of the article accurately and
    completely. It focuses on the essential information and avoids any
    redundancies and unnecessary or irrelevant content.

40-79 -- The summary includes some important information but only partially
    capture the main idea. It misses key points or includes some redundancies
    or minor details.

1-39 -- The summary fails to capture the main idea of the article and completely
    deviate from the main idea. It may focus on unimportant details or
    irrelevant content or omit critical points.

Evaluation Steps:
1. Read and understand the article.
2. Identify the article's main idea and secondary details.
3. Read the summary and judge if it captures the main idea of the article.
4. Identify if there are any secondary details or redundancy in the summary.
5. Assign a score from 1 to 100 based on the Evaluation Criteria above.

Output your detailed thought process and formal justification based on the
Evaluation Criteria, and finally output your final score in the format shown
below:

<think>
// your thought process and justification
</think>

<Final score>
Relevance: // your final score

Article:
{{ARTICLE}}

Summary:
{{SUMMARY}}
\end{lstlisting}
\end{tcolorbox}
\caption{Prompt used to evaluate summary relevance with \gemini and by the frozen base judges during RLAIF training. The article and generated summary are inserted as \texttt{\{\{ARTICLE\}\}} and \texttt{\{\{SUMMARY\}\}}.}
\label{prompt:relevance-evaluation}
\end{figure}

\begin{figure}[p]
\centering
\begin{tcolorbox}[colback=white,colframe=black,boxrule=0.4pt,arc=0pt,left=1mm,right=1mm,top=1mm,bottom=1mm]
\conversationturnlabel{User}
\begin{lstlisting}[style=conversation-turn]
You are a helpful assistant in evaluating the quality of a summary.
You will be given a news article and a summary written for that article.
Your task is to evaluate the *fluency* of the summary.

Definition of Fluency:
Fluency measures how easy the summary is to read.
All sentences in a fluent summary need to be readable and natural.
Minor grammatical, formatting, capitalization, or tokenization issues should
NOT be heavily penalized as long as they do not make the text difficult to read
or understand.

Evaluation Criteria (Fluency: 1-5)

5 -- The summary is easy to read and all the sentences are understandable and
    natural. Minor issues such as awkward wording, missing capitalization, or
    tokenization artifacts are acceptable if they do not hinder readability.

3 -- The summary is readable but some of sentences include awkward phrasing,
    inconsistent grammar, or formatting problems that reduce readability.

1 -- The summary is difficult to read. It contains frequent grammatical errors,
    broken or incomplete sentences, or severe formatting problems that
    significantly hinder understanding.

Evaluation Steps:
1. Read and understand the article.
2. Read the summary and judge whether the text is easy to read and whether the
   individual sentences are natural.
3. Assign a score from 1 to 5 based on the Evaluation Criteria above.

Only output the score. Do not include any additional explanations or text.
Use the following format:
Fluency: <score>

Article:
{{ARTICLE}}

Summary:
{{SUMMARY}}
\end{lstlisting}
\end{tcolorbox}
\caption{Prompt used to evaluate summary fluency with \gemini and by the frozen base judges during RLAIF training.}
\label{prompt:fluency-evaluation}
\end{figure}

\begin{figure}[p]
\centering
\begin{tcolorbox}[colback=white,colframe=black,boxrule=0.4pt,arc=0pt,left=1mm,right=1mm,top=1mm,bottom=1mm]
\conversationturnlabel{User}
\begin{lstlisting}[style=conversation-turn]
You are a helpful assistant in evaluating the quality of a summary.
You will be given a news article and a summary written for that article.
Your task is to evaluate the *consistency* of the summary.

Definition of Consistency:
Consistency measures how factually aligned the summary is with the article.
A consistent summary should not introduce information that contradicts the
article or contain hallucinated statements not supported by the source article.

Evaluation Criteria (Consistency: 1-5)

5 -- All statements in the summary are fully supported by the article. No
    contradictions, distortions, or hallucinated details appear.

3 -- The summary is mostly consistent but includes minor inaccuracies, unclear
    references, or slightly misleading phrasing. These issues do not
    significantly alter the meaning.

1 -- The summary contains factual errors or statements that contradict the
    article or introduce unsupported information.

Evaluation Steps:
1. Read and understand the article.
2. Read the summary and check whether each fact is supported by the article.
3. Assign a score from 1 to 5 based on the Evaluation Criteria above.

Only output the score. Do not include any additional explanations or text.
Use the following format:
Consistency: <score>

Article:
{{ARTICLE}}

Summary:
{{SUMMARY}}
\end{lstlisting}
\end{tcolorbox}
\caption{Prompt used to evaluate summary consistency with \gemini and by the frozen base judges during RLAIF training.}
\label{prompt:consistency-evaluation}
\end{figure}

\begin{figure}[p]
\centering
\begin{tcolorbox}[colback=white,colframe=black,boxrule=0.4pt,arc=0pt,left=1mm,right=1mm,top=1mm,bottom=1mm]
\conversationturnlabel{User}
\begin{lstlisting}[style=conversation-turn]
You are a helpful assistant in evaluating the quality of a summary.
You will be given a news article and a summary written for that article.
Your task is to evaluate the *coherence* of the summary.

Definition of Coherence:
Coherence measures how well the ideas in the summary fit together.
A coherent summary presents information in a logical order, with smooth
transitions between sentences. It should read as a connected, well-structured
whole.

Evaluation Criteria (Coherence: 1-5)

5 -- The summary is clearly organized and easy to follow. Sentences flow
    naturally, and ideas progress in a logical order.

3 -- The summary is somewhat coherent but has noticeable issues in flow or
    structure. Some sentences feel disconnected or out of place, yet the overall
    meaning is still clear.

1 -- The summary is hard to follow. The sentences are out of order and loosely
    connected, which makes the summary feel fragmented.

Evaluation Steps:
1. Read and understand the article.
2. Read the summary and assess whether the ideas are presented in a clear and
   logical order.
3. Assign a score from 1 to 5 based on the Evaluation Criteria above.

Only output the score. Do not include any additional explanations or text.
Use the following format:
Coherence: <score>

Article:
{{ARTICLE}}

Summary:
{{SUMMARY}}
\end{lstlisting}
\end{tcolorbox}
\caption{Prompt used to evaluate summary coherence with \gemini and by the frozen base judges during RLAIF training.}
\label{prompt:coherence-evaluation}
\end{figure}

\end{document}
\endinput